\documentclass[10pt,twocolumn,letterpaper]{article}

\usepackage[pagenumbers]{cvpr} 

\definecolor{cvprblue}{rgb}{0.21,0.49,0.74}
\usepackage[pagebackref,breaklinks,colorlinks,allcolors=cvprblue]{hyperref}
\usepackage{footmisc}
\usepackage[table]{xcolor} 
\usepackage{multirow}
\usepackage{mathtools}
\newcommand{\xcon}{x_{\text{con}}^{(c)}}
\newcommand{\mymethod}{\textsc{TEXTER}\xspace}
\usepackage{booktabs}
\usepackage{pifont}

\usepackage{xcolor}
\usepackage[most]{tcolorbox}
\usepackage{listings}
\tcbuselibrary{listings,breakable}

\newtcblisting{promptverb}[1][]{
  enhanced,
  breakable,
  colback=gray!5,
  colframe=gray!60,
  boxrule=0.4pt,
  left=1em,right=1em,
  top=0.5em,bottom=0.5em,
  listing only,
  listing options={
    basicstyle=\ttfamily\small,
    breaklines=true,
    breakatwhitespace=true,
    columns=fullflexible,
  },
}

\title{Zero-Shot Textual Explanations via Translating Decision-Critical Features}

\author{
Toshinori Yamauchi$^{1}$ \and
Hiroshi Kera$^{1,2}$ \and
Kazuhiko Kawamoto$^{1}$ \and
$^{1}$Chiba University \quad $^{2}$National Institute of Informatics\\
{\tt\small t.yamauchi@chiba-u.jp, kera@chiba-u.jp, kawa@faculty.chiba-u.jp}
}

\begin{document}
\maketitle

\begin{abstract}
Textual explanations make image classifier decisions transparent by describing the prediction rationale in natural language.  
Large vision–language models can generate captions but are designed for general visual understanding, not classifier-specific reasoning.  Existing zero-shot explanation methods align global image features with language, producing descriptions of what is visible rather than what drives the prediction.  
We propose TEXTER, which overcomes this limitation by isolating decision-critical features before alignment. 
TEXTER identifies the neurons contributing to the prediction and emphasizes the features encoded in those neurons---i.e., the decision-critical features.
It then maps these emphasized features into the CLIP feature space to retrieve textual explanations that reflect the model’s reasoning.
A sparse autoencoder further improves interpretability, particularly for Transformer architectures.  
Extensive experiments show that TEXTER provides more faithful and interpretable explanations than existing methods.
The code is available at \url{https://github.com/tttt-0814/TEXTER}.
\end{abstract}

\section{Introduction}
\label{sec:intro}
Textual explanations describe why an image classifier makes a particular prediction by revealing the visual evidence behind the decision. The explanations help users judge whether the model relies on meaningful patterns and debug the model.
Understanding what evidence classifiers use is crucial for building reliable vision systems.

\par
Recent large vision–language models (LVLMs), such as BLIP~\citep{pmlr-v162-li22n}, LLaVA~\citep{NEURIPS2023_6dcf277e}, and GPT-4V~\citep{openai2024gpt4technicalreport}, can produce descriptive captions, but these models are designed for general visual understanding rather than explaining classifier decisions.
Supervised concept-based approaches make predictions interpretable through intermediate concepts. Concept bottleneck models (CBMs)~\citep{cbn, label_free_cbn, hybrid_cbn} predict human-defined concepts before classification, requiring concept annotations and model retraining. 
Natural language explanation (NLE) models~\citep{nlx_gpt, uni_nlx} generate textual rationales from annotated explanation–label pairs. 
Both approaches reflect annotation bias and describe human-labeled concepts rather than revealing the visual evidence that drives predictions.

\begin{figure}[t]
  \centering
  \includegraphics[width=\linewidth]{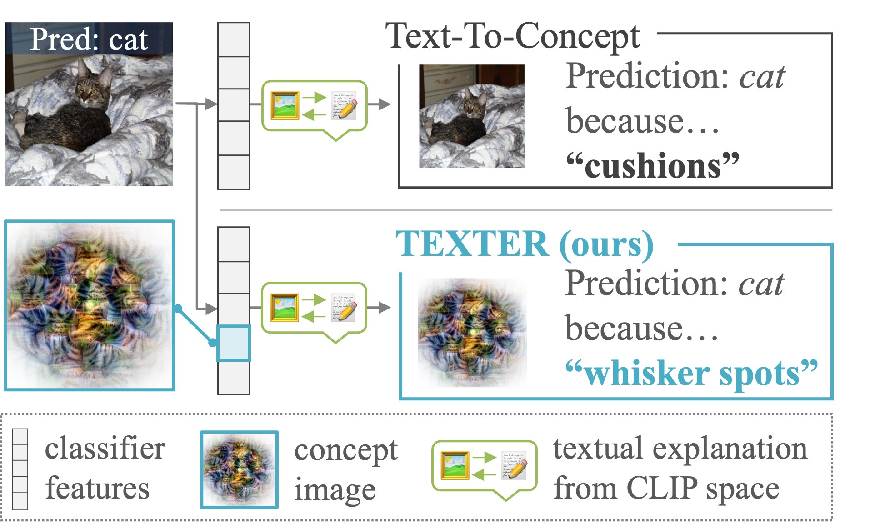}
  \caption{
  Comparison between Text-To-Concept~\citep{text_to_concept} and the proposed TEXTER for explaining a \textit{cat} prediction. 
  Text-To-Concept, which relies on global image features, produces the explanation
  ``cushions,'' describing dominant but irrelevant regions.
  In contrast, TEXTER isolates decision-critical features, such as whisker spots, through a concept image
  and translates it into the explanation ``whisker spots,'' faithfully reflecting the model's rationale.
  }
  \label{fig:intro}
\end{figure}

\par
Zero-shot alignment methods~\citep{text_to_concept, zsnle} eliminate this supervision by aligning pretrained classifier features with vision–language models such as CLIP~\citep{clip}.
These methods align entire images with language, describing \emph{what is visible} rather than \emph{what drives the prediction}.
For example, as shown in the top of \cref{fig:intro}, 
a representative zero-shot method,
Text-To-Concept~\citep{text_to_concept} outputs ``cushions'' for a \textit{cat} prediction because the method focuses on dominant visual elements instead of decision-critical features.
Therefore, a faithful explanation should align language with decision-critical features, not entire images.

To achieve this goal, 
we propose \textbf{TEXTER} (\textbf{T}extual \textbf{EX}planations via \textbf{T}ranslating d\textbf{E}cision-c\textbf{R}itical features), 
which isolates the decision-critical features before alignment.
TEXTER realizes this isolation by emphasizing the features encoded in the neurons contributing to the prediction.
Specifically, 
TEXTER first identifies the neurons that contribute to the prediction using Integrated Gradients~\citep{ig}.
The concepts encoded by these neurons are then visualized through feature visualization~\citep{maco}.
A sparse autoencoder (SAE)~\citep{topk_sae} is then applied to obtain clearer and more interpretable concept representations.
The resulting \emph{concept images} emphasize the internal visual evidence that drives the classifier's prediction.
As illustrated in the second row of \cref{fig:intro}, the concept image for a \textit{cat} prediction highlights the internal visual evidence contributing to the prediction, and translating this concept image into natural language yields the explanation ``whisker spots.''
TEXTER performs this translation by mapping concept images into the CLIP feature space to retrieve textual descriptions that reflect the classifier’s reasoning.  
Focusing on decision-critical features rather than global representations, TEXTER differs from previous zero-shot approaches and yields more faithful textual explanations.

Extensive experiments demonstrate that TEXTER produces faithful and interpretable explanations across both CNN and Transformer architectures.  
Quantitative analyses show that the concept images capture decision-critical features, while qualitative results confirm that the provided textual explanations reflect the model’s actual reasoning.

\par
Our contributions are summarized as follows.
\begin{itemize}
  \item {We propose {TEXTER}, 
  which translates the decision-critical features of classifiers into the CLIP feature space to provide textual explanations.}
  \item {We demonstrate generalizability across both CNN and Transformer architectures by incorporating the SAE to isolate decision-critical features, enabling TEXTER to perform effectively across these model types.}
  \item {Extensive experiments show that TEXTER provides explanations that faithfully reflect the rationale behind the predictions.
  Both quantitative and qualitative evaluations confirm effectiveness.  
  }
\end{itemize}

\section{Related work}
\label{sec:related_work}
Providing textual explanations for image classification models has been widely explored, ranging from concept-based reasoning to large vision–language models. 
Here, we discuss representative approaches, 
including concept bottleneck models and natural language explanation models, 
and describe how our study differs from them. 
Finally, we explain recent zero-shot approaches and their limitations.

\paragraph{Concept bottleneck models (CBMs).}
Early concept-based interpretability methods, such as Network Dissection~\citep{8099837} and TCAV~\citep{tcav}, evaluated how neurons align with human-understandable concepts.
Later works introduced Concept Bottleneck Models (CBMs)~\citep{cbn, label_free_cbn, labo, incremental_cbn, language_cbn},
which predict human-defined concepts before the final label, 
improving interpretability but requiring concept-labeled data.
Recent studies aim to reduce this annotation cost---for instance, Label-Free CBM~\citep{label_free_cbn} leverages CLIP for automatic concept labeling, and Hybrid CBM~\citep{hybrid_cbn} augments predefined concepts with LLM-generated ones.
Unlike these studies, 
our work targets classification models trained solely on images and aims to provide textual explanations without retraining the original model,
thereby broadening the applicability of concept-based interpretability to existing pretrained classifiers.

\begin{figure*}[t]
  \centering
  \includegraphics[width=\linewidth]{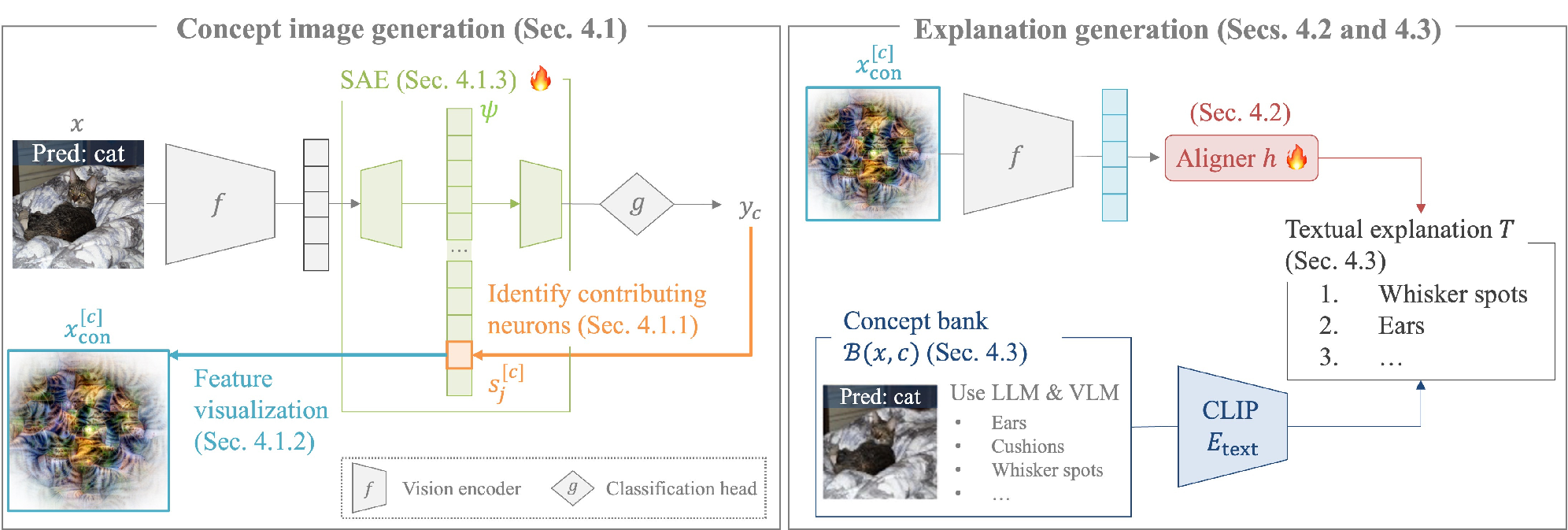}
  \caption{Overview of the proposed TEXTER.
    The left part illustrates the concept image generation process described in \cref{sec:concept_image_generation}, while the right part shows the explanation generation process, in which each component is detailed in \cref{sec:alignment,sec:textual_explanation_generation}.
    Only the SAE modules and the aligner $h$ are trainable, using a subset of the classifier’s training dataset.
    Note that these trainable modules require no additional annotations and are not involved in the inference process of the classifier, and therefore do not affect the original accuracy.
}
  \label{fig:method}
\end{figure*}

\paragraph{Natural language explanation (NLE) models.}
NLE models are trained to describe why a vision or vision–language model makes a prediction~\citep{generate_visual_explanation, park_nle, nlx_gpt, sharma, debil}.
They typically combine a task model (e.g., a classifier) with a language model such as GPT-2~\citep{gpt2} to generate explanations, 
as in NLX-GPT~\citep{nlx_gpt}, which unifies the two components, 
and Uni-NLX~\citep{uni_nlx}, which extends the framework to multiple vision–language tasks.
However, these models depend on paired prediction–explanation annotations and often reproduce annotation bias rather than the model’s true reasoning~\citep{zsnle}.
In contrast, our approach directly generates textual explanations for classification models in a zero-shot setting.
Recent large vision–language models (LVLMs), such as BLIP~\citep{pmlr-v162-li22n}, LLaVA~\citep{NEURIPS2023_6dcf277e}, and GPT-4V~\citep{openai2024gpt4technicalreport}, 
have demonstrated remarkable ability in generating open-ended visual descriptions and reasoning about images. 
These models, however, are trained for generic vision–language understanding, 
whereas our work focuses on explaining the decision process of a specific classifier.

\paragraph{Zero-shot textual explanations.}
Recent work has explored zero-shot methods for generating textual explanations.
While several studies leverage vision–language models~\citep{menon, shtedritski, zs-a2t, interpreting_clip}, approaches tailored to image classifiers remain limited.
Text-To-Concept~\citep{text_to_concept} addresses this by linearly aligning the classifier feature space with CLIP~\citep{clip}, 
enabling direct comparison between visual features and text embeddings.
Its zero-shot classification accuracy demonstrates the effectiveness of this alignment, eliminating the need for manual concept annotation required by methods such as TCAV~\citep{tcav, fel2023, zhang}.
A related approach, 
ZSNLE~\citep{zsnle} trains a lightweight multi-layer perceptron to align the CLIP text-encoder space with the classifier’s weight space.
Unlike Text-To-Concept, which projects classifier features into the vision–language space, ZSNLE performs the inverse mapping.
Another work~\cite{beyond_clip} further develops this CLIP-alignment framework by decomposing representations into component-wise contributions aligned to textual attributes.
However, these methods rely on global image features rather than those directly responsible for the model’s decision,
which motivates our proposed method.
Overall, while concept-based and vision–language models have advanced textual interpretability, 
existing methods either depend on annotated concept labels or fail to isolate the neurons most responsible for a prediction. 
Our work addresses this gap by combining neuron-level concept visualization with zero-shot textual grounding.

\section{Problem setup}
\label{sec:setup}
We set up our problem.
Let $\mathcal{X}$ denote the set of input images and let the classifier be $\mathcal{F}=g\circ f:\mathcal{X}\to\mathcal{Y}$,
where $f:\mathcal{X}\to Z_f$ is a vision encoder mapping an input $x\in\mathcal{X}$ to a feature vector $f(x)\in Z_f$,
and $g:Z_f\to\mathcal{Y}$ is a classification head.
We take $\mathcal{Y}=\mathbb{R}^{C}$ to be the logit space for a $C$-class problem, and write $y\in\mathcal{Y}$ for a logit vector.
The classifier is trained solely on image domain, 
and no language modeling is associated. 
For such $\mathcal{F}$, 
we want to explain its decision (i.e., prediction) in natural language.

\par
To this end, we have two challenges. 
The first is how to associate the latent feature space of the target classifier (i.e., $Z_f$) with some language domain.
We introduce a VLM that consists of an image encoder $E_{\text{img}}: \mathcal{X} \to Z_{E_{\text{img}}}$ and a text encoder $E_{\text{text}}: \mathcal{T} \to Z_{E_{\text{text}}}$, where $\mathcal{T}$ denotes the set of text descriptions.
Since the target classifier and the VLM do not share a common feature space, 
our first goal is to align $Z_f$ with $Z_{E_{\text{img}}}$.
Once the alignment is completed,
the visual and linguistic features can be associated within the VLM feature space.

\par
The second challenge is to isolate decision-critical information from the feature space of the target classifier.
A direct alignment of the global image feature to the VLM space, 
as in prior work~\citep{text_to_concept,zsnle}, 
captures only dominant visual attributes rather than explaining the model’s decision (see \cref{fig:intro}).
Therefore, 
our second goal is to find a projection $\varphi$ that isolates the decision-critical feature $z^{(c)}$ for the predicted class $c$ from the global image feature $f(x)$.

\section{Method}
\label{sec:method}

We present TEXTER,
a zero-shot framework that explains the rationale behind the prediction in natural language.
Figure~\ref{fig:method} illustrates the overview of TEXTER.
It has three stages, concept image generation, Vision--Language space alignment, and textual explanation generation.
Here, we provide an overview of each stage, 
followed by detailed explanations in the subsequent sections.

\paragraph{Concept image generation (\cref{sec:concept_image_generation}).}
As discussed in \cref{sec:setup},
we need to find a projection from the feature vector $f(x)$, 
which captures the global image feature, 
to the decision-critical feature $z^{(c)}$ for the predicted class $c$:
\begin{align}
f(x) \quad \overset{\varphi}{\longmapsto}\quad z^{(c)}.
\end{align}
The concept images generated through feature visualization~\citep{maco} serve this purpose.
Specifically, 
the concept image $\xcon$ associated with $x$ is designed to emphasize the decision-critical feature of $x$ for the predicted class $c$ in the latent space,
and we ragard $f(\xcon)$ as the decision-critical feature $z^{(c)}$.
Therefore, 
the process $f(x) \mapsto \xcon \mapsto f(\xcon)$ can be regarded as a realization of $\varphi$.

\paragraph{Vision--Language space alignment (\cref{sec:alignment}).}
The next step aligns the feature space $Z_f$ with $Z_{E_{\text{img}}}$.
This alignment is achieved by training an affine layer $h$ to bring the two feature spaces closer.
After alignment, 
the text space $Z_{E_{\text{text}}}$ becomes associated with the latent space of the target classifier.

\paragraph{Textual explanation generation (\cref{sec:textual_explanation_generation}).} 
Explanations are generated  by comparing the aligned classifier features 
for the concept image $\xcon$ with the textual features of the VLM.
Since an infinite set of textual descriptions cannot be compared directly, candidates descriptions are retrieved  from a concept bank $\mathcal{B}(x,c)$, which contains a set of plausible concept descriptions.
The top-$k_{\text{con}}$ descriptions with the highest similarity scores are then selected as the textual explanations.
Unlike prior works that use global image features $f(x)$ for comparison, 
TEXTER compares $f(\xcon)$---the decision-critical features $z^{(c)}$---thereby producing explanations that reflect the model’s decision.

\subsection{Concept image generation}
\label{sec:concept_image_generation}

The mapping $\varphi$, 
which isolates the decision-critical features $z^{(c)}$ for the predicted class $c$, is realized through the concept image $\xcon$.
The key insight is that the most influential neurons encode the features underlying the prediction~\citep{milan,clip_dissect}, 
and the concept image is designed to emphasize these decision-critical features.
The generation process involves three steps: identifying the contributing neurons, visualizing their concepts, and applying a Sparse Autoencoder (SAE) to improve interpretability.

\subsubsection{Identifying the contributing neurons}
\label{sec:identify_neurons}

A contribution score is computed for each neuron, and the top-scoring neurons are selected.
Let $z$ be the feature vector at an arbitrary layer of the classifier for an input image $x$ (i.e., $z = f^{(\ell)}(x)$).
The contribution score $s_j^{(c)}$ for the $j-$th neuron $z_j$ is computed using integrated gradients~\citep{ig}:
\begin{align}
s_j^{(c)}
&=\big(z_j - z'_j\big)
\sum_{m=1}^{M}
\frac{\partial F_c\big(z' + \tfrac{m}{M}(z - z')\big)}{\partial z_j}
\cdot \frac{1}{M},
\label{eq:ig_discrete}
\end{align}
where the baseline $z'$ serves as a reference feature vector from which the integrated gradients accumulate contributions toward the final prediction,
and $M$ denotes the number of steps\footnote{The baseline $z'$ is set to the zero vector and $M = 100$ by default.}.
Here, $F_c$ denotes the function that maps a feature vector to the class-$c$ logit $y_c$, and it is evaluated at $\hat{z} := z' + \tfrac{m}{M}(z - z')$, i.e., $y_c=F_c(\hat{z})$.
A higher score indicates that the corresponding neuron contributes more to the prediction~\citep{ig}.

\par
After computing the scores for all neurons in $z$,
the neurons are sorted in descending order of $s_j^{(c)}$,
and the top-$k_{\text{neu}}$ indices are selected:
\begin{align}
U^{(c)} = \{u_1, \ldots, u_{k_{\text{neu}}}\}.
\label{eq:topk_neurons}
\end{align}
The neurons indexed by $U^{(c)}$ are those contributing most to the class-$c$ prediction.

\subsubsection{Visualizing concepts in contributing neurons}
\label{sec:vis_concepts}

The concepts encoded by the identified neurons are visualized to obtain the concept image $\xcon$.
We employ feature visualization~\citep{fv,nguyen16,nguyen17,maco,vital},
a general approach for understanding internal representations of neural networks.
Feature visualization optimizes an input $x' \in \mathcal{X}$ to maximize a criterion $\mathcal{L}(x') \in \mathbb{R}$,
formulated as:
\begin{align}
\xcon= 
\underset{\substack{x' \in \mathcal{X}}}{\operatorname{argmax}}\,
\mathcal{L}(x') - \lambda\mathcal{R}(x'),
\label{eq:fv_general}
\end{align}
where $\lambda$ balances the $\mathcal{L}$ and the regularizer $\mathcal{R}$.
We adopt magnitude-constrained optimization (MACO)~\citep{maco},
which optimizes the phase spectrum while keeping the magnitude constant in the Fourier space,
ensuring that the generated image remains within the natural image distribution.
The criterion $\mathcal{L}$ is defined as the sum of the activations of the identified neurons:
\begin{align}
\mathcal{L}(x') = \sum_{j \in U^{(c)}} [f^{(\ell)}(x')]_j,
\label{eq:criterion}
\end{align}
where $[f^{(\ell)}(x')]_j$ denotes the $j$-th element of $f^{(\ell)}(x')$.
From \cref{eq:fv_general,eq:criterion},
the concept image $\xcon$ represents the features encoded by the neurons contributing to the prediction.

\subsubsection{SAE for interpretable concept representations}
\label{sec:sae}

We employ SAE~\citep{fel2023a,bhalla24} to obtain more interpretable concept representations.
Empirically, we find that the SAE is not essential for CNNs but crucial for Transformers, which recognize objects more compositionally~\citep{vit_cnn1,vit_cnn2,decision_making}, making their entangled feature space less effective for isolating decision-critical factors~\citep{usae}.
As shown in previous studies, SAEs factorize DNN features into sparse, axis-aligned units, enhancing interpretability.
See \cref{sec:eval_concept_image} for detailed analysis.

\par
We adopt the TopK SAE~\citep{topk_sae,usae},
which learns an encoder $\Psi(\cdot)$ that maps $f(x)$ to a sparse representation:
\begin{align}
\Psi(f(x)) = \operatorname{TopK}(W_{\text{enc}} (f(x) - b_{\text{pre}})),
\label{eq:topk_sae}
\end{align}
where $W_{\text{enc}}$ and $b_{\text{pre}}$ are trainable weights.
The operator $\operatorname{TopK}(\cdot)$ keeps the $K$ largest entries of a vector and sets the others to zero.
Training minimizes the reconstruction loss:
\begin{align}
L_{\text{SAE}} = \left||f(x) - W_{\text{dec}}\Psi(f(x)) \right||_2^2,
\label{eq:loss_sae}
\end{align}
where $W_{\text{dec}}$ denotes the decoder weights.
Details of the SAE configuration are provided in Appendix~\ref{sec:sae_config}.

\par
TEXTER identifies the contributing neurons in its sparse representation and generates concept images for them.
Accordingly, by regarding the SAE as an additional module of the classifier, in \cref{eq:ig_discrete},
$z = \Psi(f(x))$,
and $y_c=F_c(\hat{z}) = [g(W_{\text{dec}}\hat{z})]_c$.
In \cref{eq:criterion},
$f^{(\ell)}(x')$ is replaced with $\Psi(f(x'))$.

\subsection{Vision-Language space alignment}
\label{sec:alignment}
We use CLIP~\cite{clip} as the VLM and align the classifier feature space with the CLIP vision feature space.
To achieve this, 
following Text-To-Concept~\citep{text_to_concept}, an affine aligner $h(f(x)) = Wf(x) + b$ is trained by minimizing the alignment loss:
\begin{align}
\min_{W,b}
\frac{1}{|\mathcal{D}_{\text{train}}|}
\sum_{x \in \mathcal{D}_{\text{train}}}
\left\| W f(x)+b-E_{\text{img}}(x) \right\|_2^2,
\label{eq:train_h}
\end{align}
where $\mathcal{D}_{\text{train}}$ denotes the training dataset, and $W$ and $b$ are trainable weights.
After training is completed, 
the aligned features lie in the joint vision–language feature space.

\subsection{Textual explanation generation}
\label{sec:textual_explanation_generation}

After training, 
the aligner $h$ maps the features of the concept image into the CLIP feature space.
To obtain textual explanations,
similarities between the aligned feature $h(f(\xcon))$ for the concept image and the text embeddings $E_{\text{text}}(t)$
of all candidate descriptions $t \in \mathcal{B}(x,c)$ (see the next paragraph)
are computed using cosine similarity.
\begin{align}
\operatorname{sim}(h(f(\xcon)), E_{\text{text}}(t_i)),
\quad t_i \in \mathcal{B}(x,c),
\label{eq:sim_texter}
\end{align}
where $\operatorname{sim}(\cdot,\cdot)$ denotes cosine similarity.
The top-$k_{\text{con}}$ descriptions with the highest similarity scores are then selected as the textual explanations,
$T = \{t_1, \ldots, t_{k_{\text{con}}}\}$.

\paragraph{Construction of concept bank.} 
The concept bank $\mathcal{B}(x,c)$ is a set of plausible concept descriptions,
collected to include those likely to be relevant to the prediction $c$ of the input image $x$.
Given an input image $x$ and its predicted class $c$,
TEXTER constructs concept bank $\mathcal{B}(x,c)$ by leveraging both a large language model (LLM) and a vision-language model (VLM).
Inspired by prior studies~\citep{label_free_cbn,labo,menon},
an LLM is prompted to generate generic descriptions characterizing the predicted class $c$.
Complementarily,
a VLM is prompted with the input image $x$ and class $c$ to generate visually grounded descriptions.
We use
\texttt{GPT\mbox{-}3.5\mbox{-}turbo}~\citep{gpt3.5} as the LLM and
\texttt{Qwen2.5\mbox{-}VL\mbox{-}7B\mbox{-}Instruct}~\citep{Qwen-VL,Qwen2VL,qwen2.5-VL} as the VLM.
Further details are provided in Appendix~\ref{sec:detail_concept_bank}.

\begin{table}[t]
\caption{Evaluation of concept image validity (higher is better).
A checkmark (\checkmark) indicates that the concept images are generated from the SAE feature vectors, whereas a dash (–) indicates that they are generated from the feature vectors of the classifier’s vision encoder $f$.}
\centering
\label{tab:concept_eval_grouped}
\small
\begin{tabular}{c c r r r r}
\toprule
\textbf{Model} & \textbf{SAE} & \textbf{Acc$_1$} & \textbf{Acc$_5$} & \textbf{$\text{R}_{\text{conf}}$} & \textbf{Cos} \\
\midrule
ResNet-18      & - & 0.92 & 1.00 & 1.38 & 0.72 \\
               &  \checkmark & 0.80 & 0.96 & 1.16 & 0.66 \\
\midrule
ResNet-50      & - & 0.92 & 1.00 & 1.19 & 0.74 \\
               &  \checkmark & 0.83 & 0.97 & 1.05 & 0.70 \\
\midrule               
DINO ResNet-50 & - & 0.79 & 0.94 & 1.07 & 0.63 \\
               &  \checkmark & 0.48 & 0.80 & 0.61 & 0.63 \\
\midrule
ViT            & - & 0.11 & 0.17 & 0.06 & 0.11 \\
               &  \checkmark & 0.99 & 1.00 & 1.04 & 0.44 \\
\midrule
DINO ViT-S/8   & - & 0.08 & 0.15 & 0.06 & 0.34 \\
               &  \checkmark & 0.89 & 0.96 & 0.91 & 0.61 \\
\bottomrule
\end{tabular}
\end{table}

\section{Experiments}
\label{sec:experiments}

We evaluate TEXTER on both CNN- and Transformer-based classifiers to validate that concept images capture decision-critical features rather than dominant image content.
We first assess whether concept images preserve the information used in predictions (\cref{sec:eval_concept_image}).
We then qualitatively and quantitatively compare generated explanations with existing methods in multi-label classification setting (\cref{sec:multilabel})
and qualitatively analyze the reasoning patterns TEXTER reveals across architectures (\cref{sec:qualitative_evaluations}).
Finally, we demonstrate TEXTER's effectiveness in class-wise explanation scenarios (\cref{sec:class_wise_explanations}).

\noindent\textbf{Models and setup.}
We use ResNet-18, ResNet-50~\citep{resnet}, and DINO ResNet-50~\citep{dino} for CNNs, 
and ViT~\citep{vit} and DINO ViT-S/8~\citep{dino} for Transformers.
The aligner $h$ maps the classifier features into the CLIP feature space using the ViT-B/16 vision encoder.
We adopt the same training data and protocol as \citep{text_to_concept}, using 20\,\% of the ImageNet-1K training dataset~\citep{imagenet} for $\mathcal{D}_{\text{train}}$.
For aligners with publicly released weights, 
we use the official checkpoints provided in the authors’ GitHub repository.\footnote{https://github.com/k1rezaei/Text-to-concept}
All other training settings for the aligner follow the original configuration in \citep{text_to_concept}.
In \cref{eq:topk_neurons}, we set $k_\text{neu} = 6$.
For the concept bank, unless otherwise specified, we obtain 100 descriptions from the LLM and 30 from the VLM, resulting in 130 descriptions in $\mathcal{B}(x,c)$ (i.e., $|\mathcal{B}(x,c)| = 130$).
Additional implementation details are provided in Appendix~\ref{sec:other_implementation_details}.
Unless otherwise specified, the following experiments use the ImageNet-1K dataset.

\subsection{Validity assessment of concept images}
\label{sec:eval_concept_image}
TEXTER generates explanations from concept images $\xcon$, which are produced from the SAE feature vectors, rather than from the original images.
Therefore, we assess whether the concept images preserve the features that drove the original predictions of the corresponding classifiers.

\noindent\textbf{Metrics.}
We use three metrics.
\begin{itemize}[leftmargin=1.0em]
\item {Accuracy (Acc)}: 
Top-1 and Top-5 accuracy for the original predicted label $c$ when classifying concept images.
\item {Confidence ratio ($\text{R}_\text{conf}$)}:
Ratio of model confidence (softmax probability) for class $c$ on the concept image to that on the original image.
\item {Cosine similarity (Cos)}:
Cosine similarity between the logit vectors of the original and concept images
$\cos(g(f(x)), g(f(\xcon)))$.
\end{itemize}
Higher values across all metrics indicate better preservation of decision-critical information.

\noindent\textbf{Results.}
Table~\ref{tab:concept_eval_grouped} 
summarizes results on 1,000 randomly selected ImageNet-1K test images 
(200 classes, 5 images per class).
We compare the concept images generated directly from the classifier’s vision encoder $f$ 
(i.e., without applying the SAE).

\par
Without the SAE, concept image validity remains high for CNN models, whereas Transformer models obtain very low scores.
This difference is consistent with
the compositional and entangled nature of Transformer feature spaces~\citep{vit_cnn1,vit_cnn2,decision_making,usae}, 
which makes it difficult to isolate decision-critical features from the raw representations.
Applying the SAE addresses this issue and leads to large improvements for Transformer models.
For CNNs, the sparsity imposed by the SAE can introduce small decreases (e.g., DINO ResNet-50), but the resulting concept images remain largely comparable.

Overall, TEXTER with the SAE generates concept images that 
reliably capture decision-critical features across both CNN and Transformer architectures.

\begin{figure}[t]
  \centering
  \includegraphics[scale=0.5]{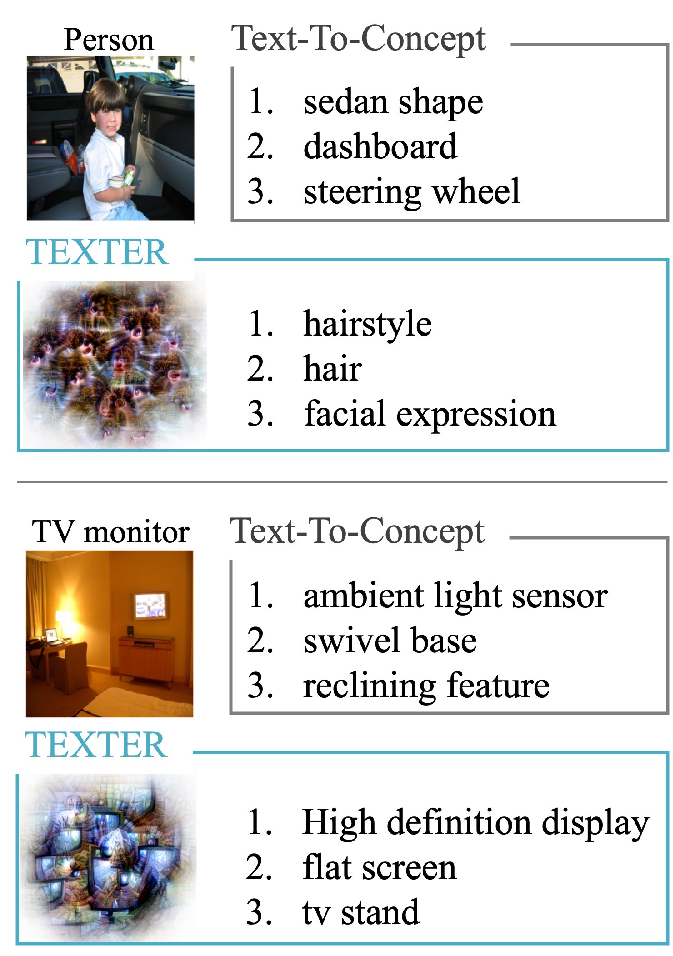}
  \caption{Comparison of provided explanations between Text-To-Concept and the proposed method.
    The figure presents two cases: the prediction of \textit{person} (top) and the prediction of \textit{TV monitor} (bottom).
    For each result generated by TEXTER, the corresponding concept image is displayed.}
  \label{fig:multilabel}
\end{figure}

\begin{table*}[t]
\centering
\caption{Quantitative evaluation using semantics-based metrics. 
Each metric is computed between the explanations provided by each method and the concept images generated by the proposed method.
Best results in bold.
}
\label{tab:results_for_concept_pascal}
\begin{tabular}{l l c c c c}
\toprule
\textbf{Model} & \textbf{Method} & \textbf{CLIP-Score} $\uparrow$ & \textbf{LPIPS (A)} $\downarrow$ & \textbf{LPIPS (S)} $\downarrow$ & \textbf{FS} $\uparrow$ \\
\midrule
\multirow{3}{*}{ResNet-18}
  & Random                      & 0.2065 & 0.7591 & 0.7045 & 0.5730 \\
  & Text-To-Concept             & 0.2086 & 0.7591 & 0.7055 & 0.5703 \\
  & \cellcolor{gray!12}\mymethod  
                                & \cellcolor{gray!12}\textbf{0.2155} & \cellcolor{gray!12}\textbf{0.7481} & \cellcolor{gray!12}\textbf{0.6938} & \cellcolor{gray!12}\textbf{0.5994} \\
\midrule
\multirow{3}{*}{ResNet-50}
  & Random                      & 0.2050 & 0.7775 & 0.7041 & 0.5874 \\
  & Text-To-Concept             & 0.2073 & 0.7698 & 0.7017 & 0.6032 \\
  & \cellcolor{gray!12}\mymethod  
                                & \cellcolor{gray!12}\textbf{0.2271} & \cellcolor{gray!12}\textbf{0.7609} & \cellcolor{gray!12}\textbf{0.6972} & \cellcolor{gray!12}\textbf{0.6378} \\
\midrule
\multirow{3}{*}{DINO ResNet-50}
  & Random                      & 0.2065 & 0.7579 & 0.6957 & 0.5042 \\
  & Text-To-Concept             & 0.2172 & 0.7485 & 0.6938 & 0.5220 \\
  & \cellcolor{gray!12}\mymethod
                                & \cellcolor{gray!12}\textbf{0.2340} & \cellcolor{gray!12}\textbf{0.7385} & \cellcolor{gray!12}\textbf{0.6882} & \cellcolor{gray!12}\textbf{0.5512} \\
\midrule
\multirow{3}{*}{ViT}
  & Random                      & 0.2127 & 0.7804 & 0.6971 & 0.1329 \\
  & Text-To-Concept             & 0.2187 & 0.7909 & 0.7006 & 0.1458 \\
  & \cellcolor{gray!12}\mymethod
                                & \cellcolor{gray!12}\textbf{0.2283} & \cellcolor{gray!12}\textbf{0.7727} & \cellcolor{gray!12}\textbf{0.6952} & \cellcolor{gray!12}\textbf{0.1526} \\
\midrule
\multirow{3}{*}{DINO ViT-S/8}
  & Random                      & 0.2182 & 0.7648 & 0.6943 & 0.3686 \\
  & Text-To-Concept             & 0.2224 & 0.7592 & 0.6946 & 0.3933 \\
  & \cellcolor{gray!12}\mymethod
                                & \cellcolor{gray!12}\textbf{0.2334} & \cellcolor{gray!12}\textbf{0.7539} & \cellcolor{gray!12}\textbf{0.6937} & \cellcolor{gray!12}\textbf{0.4082} \\
\bottomrule
\end{tabular}
\end{table*}

\begin{figure*}[t]
  \centering
  \includegraphics[width=\linewidth]{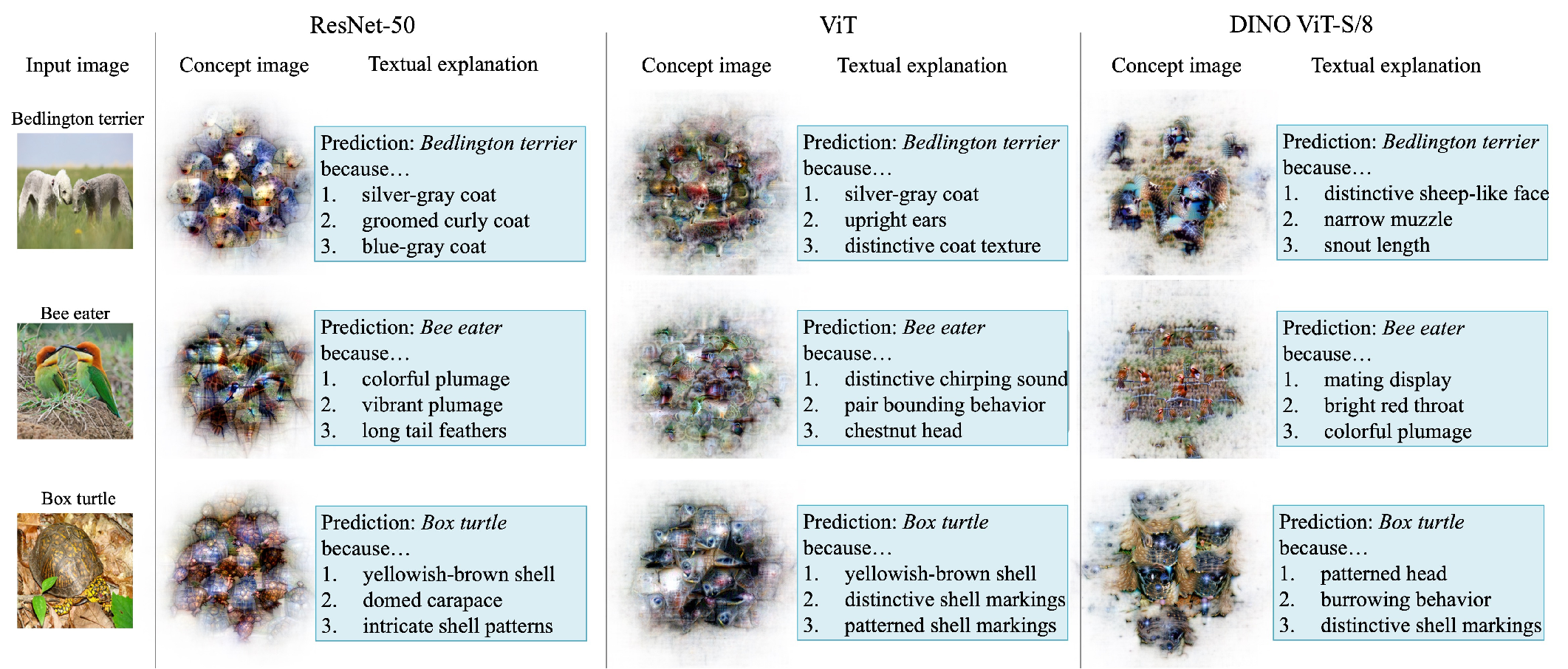}
  \caption{Qualitative comparison of the textual explanations and concept images generated by the proposed method for the same prediction across ResNet-50, ViT, and DINO ViT-S/8. Each row corresponds to one input image and its predicted class.}
  \label{fig:comp_per_model}
\end{figure*}

\subsection{Evaluation of textual explanations}
\label{sec:multilabel}
Existing methods that rely on global image features often highlight dominant image contents rather than the actual rationale behind predictions. 
This limitation becomes more apparent in images containing multiple objects. 
To demonstrate this limitation and the effectiveness of TEXTER, 
we conduct experiments in a multi-label classification setting.

We use each pre-trained classifier and fine-tune only its classification layer on the PASCAL VOC dataset~\citep{voc},
where each image may contain multiple object instances.
The model produces class-wise predictions via sigmoid activation and outputs multiple classes whose predicted probabilities exceed a threshold of 0.3.
To accommodate this setting, 
the concept bank contains the union of concept sets from the predicted classes (130 descriptions per class).
In the following, we discuss both the qualitative and quantitative results.

\subsubsection{Qualitative results}
Figure~\ref{fig:multilabel} compares explanations provided by Text-To-Concept and TEXTER.
Text-To-Concept describes visually dominant features regardless of predicted classes.
For \textit{person} (top), 
Text-To-Concept focuses on background elements (e.g., sedan shape)
rather than person-specific features.
For \textit{TV monitor} (bottom), 
Text-To-Concept highlights furniture features (e.g., light sensor) instead of monitor-specific features.
This occurs because Text-To-Concept relies on global image features.
TEXTER provides more appropriate explanations by extracting
features from concept images.
For \textit{person},
TEXTER identifies hairstyle, hair, and facial expression; 
\textit{TV monitor},
TEXTER identifies high-definition display, flat screen, and tv stand.
These results demonstrate that concept images help isolate 
the decision-critical features that drive model predictions.

\subsubsection{Quantitative results}
\label{sec:quantitative_evaluations}
We quantitatively evaluate textual explanations using semantics-based metrics, following~\citep{zsnle}.
These metrics measure how well textual explanations semantically align with the images.
Prior work~\cite{zsnle} compares explanations with the original images; however, such evaluations assess only how well the explanations describe the image rather than the reasoning behind the prediction.
Therefore, we use concept images as comparison targets, as they highlight decision-critical features rather than the visual content itself (as discussed in \cref{sec:eval_concept_image}).
We also conduct evaluations on ImageNet-1K and, 
for completeness, include a comparison with the original images (see Appendix~\ref{sec:additional_semantic_eval}).

\noindent\textbf{Metrics.}
Given textual explanations $T$ and concept images $\xcon$, 
we evaluate semantic consistency using the following metrics.
Further details are provided in Appendix~\ref{sec:experimental_settings}.

\begin{itemize}
\item{CLIP-Score}~\citep{clip_score}: Text-image semantic similarity between $\xcon$ and $t\in T$ 
in CLIP embedding space.
\item{LPIPS}~\citep{lpips}: 
Perceptual similarity.
We generate the image $x_g$ from $T$ using Stable 
Diffusion~\citep{stable_diffusion} 
and compute LPIPS between $\xcon$ and $x_g$
using pretrained LPIPS models\footnote{https://github.com/richzhang/PerceptualSimilarity}
based on AlexNet (A)~\citep{alexnet} and SqueezeNet (S)~\citep{squeezenet}.
\item{Feature Similarity (FS).}
Cosine similarity between classifier features, 
$\cos\big(f(\xcon), f(x_g)\big)$.
\end{itemize}

\noindent\textbf{Baselines.}
We compare TEXTER with the following baselines. 
\begin{itemize}
\item \textbf{Random}, 
which randomly selects textual explanations from the concept bank.
\item \textbf{Text-To-Concept}~\citep{text_to_concept}, 
which generates explanations following \cref{eq:sim_texter} but computes similarity using the original image $x$ instead of the concept image $\xcon$.
\end{itemize}
In both Text-To-Concept and TEXTER, 
the top-3 textual explanations are used ($k_{\text{con}}=3$),
whereas Random selects three at random.

\noindent\textbf{Results.}
The results are shown in \cref{tab:results_for_concept_pascal}.
All results are computed on 100 randomly selected images from the PASCAL VOC test set, using the top-scoring class for each image.
Note that all methods are evaluated using concept images generated by TEXTER.
TEXTER obtains higher scores than Random and Text-to-Concept across all metrics and models.
These results indicate that TEXTER provides explanations with better semantic alignment with the concept images than the baselines.
We additionally evaluate faithfulness via a text-to-region insertion/deletion test; details are provided in Appendix~\ref{sec:faithfulness_eval}.

\subsection{Qualitative comparison across models}
\label{sec:qualitative_evaluations}
Figure~\ref{fig:comp_per_model} compares the textual explanations and concept images generated by TEXTER for the same prediction on the ImageNet dataset across ResNet-50, ViT, and DINO ViT-S/8. 
The results reveal model-dependent tendencies in which features drive predictions.

\par
For Bedlington terrier (first row), 
ResNet-50 and ViT focus on texture attributes such as the silver-gray coat, 
whereas DINO ViT-S/8 attends to structural facial features such as the sheep-like face and narrow muzzle.
For Bee eater (second row), 
ResNet-50 highlights partial visual features such as colorful plumage, 
whereas ViT and DINO ViT-S/8 retrieve concepts related to interactions between two birds, such as pair bonding behavior and mating display.
For Box turtle (third row), 
ResNet-50 and ViT primarily describe shell appearance,
whereas DINO ViT-S/8 also captures behavioral context through concepts like burrowing behavior.

\par
These observations show that Transformer-based models, particularly self-supervised ones like DINO,
incorporate relational and contextual information alongside appearance features~\citep{vit_cnn1,vit_cnn2,decision_making}.
Additional results are provided in Appendix~\ref{sec:additional_comparison}.

\begin{figure}[t]
  \centering
  \includegraphics[width=0.8\linewidth]{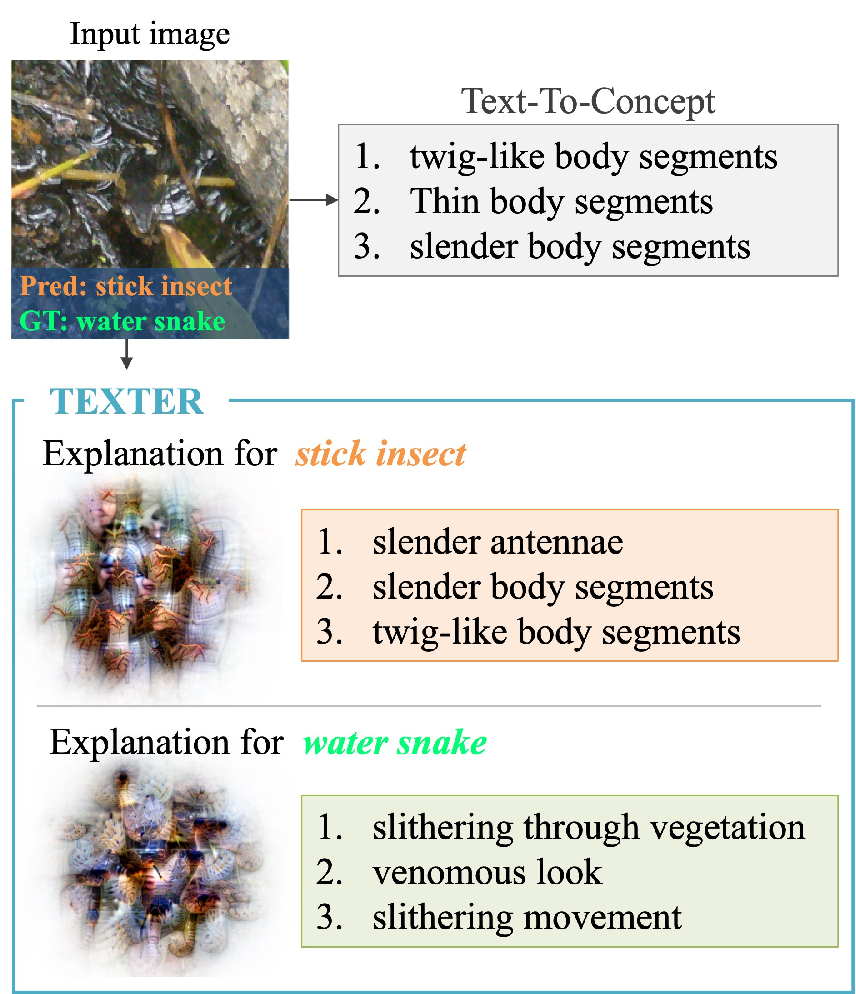}
  \caption{Comparison of the provided explanations between Text-To-Concept and the proposed method for an input whose ground-truth label is \textit{water snake} but is misclassified as \textit{stick insect}. For the proposed method, concept images targeting each class are shown. All explanations are provided from a shared concept bank constructed as the union of those for \textit{stick insect} and \textit{water snake}.}
  \label{fig:class_wise_explanation}
\end{figure}

\subsection{Class-wise explanations}
\label{sec:class_wise_explanations}
Existing zero-shot approaches typically rely on global image--text similarity and therefore do not explicitly support class-conditioned analysis without providing an external class query.
In contrast, TEXTER retrieves explanations conditioned on an arbitrary target class, enabling class-wise analysis in multi-class settings. This is particularly important when multiple classes share overlapping visual cues, where background structures may resemble features of competing classes.

\par
Figure~\ref{fig:class_wise_explanation} illustrates an example where the ground-truth label is \textit{water snake} but the model predicts \textit{stick insect}. 
Text-To-Concept produces only class-agnostic descriptions,
whereas TEXTER generates explanations for both classes.
With the target class set to \textit{stick insect}, 
TEXTER produces explanations such as slender antennae and slender body segments, 
indicating that twig-like background fragments are misinterpreted as insect limbs. 
With the target class set to the ground-truth \textit{water snake}, 
TEXTER produces explanations such as slithering through vegetation and venomous look, 
suggesting that cues relevant to the ground-truth class remain present despite the incorrect prediction.

\par
This class-conditioned analysis helps clarify the reasoning behind both the predicted class and the ground-truth class within a single image,
leading to a deeper understanding of the model's behavior.

\section{Conclusion}
\label{sec:conclusion}
We propose TEXTER, 
a zero-shot framework that explains model predictions in natural language.
TEXTER isolates decision-critical features by generating concept images and produces class-specific textual explanations by aligning concept image features with the CLIP feature space.
Incorporating a Sparse Autoencoder (SAE) yields more interpretable concept representations and improves explanation faithfulness across both CNN- and Transformer-based models.
Experimental results demonstrate that TEXTER captures decision-critical features more faithfully
than the existing methods that rely on global image features.
This work takes a step toward interpretable vision models that explain what drives model predictions in natural language more effectively.

\section*{Acknowledgments}
This work was supported by JSPS KAKENHI Grant Number JP23K24914,
JST PRESTO Grant Number JPMJPR24K4,
JST BOOST Program Grant Number JPMJBY24C6,
and ROIS NII Open Collaborative Research 261S07-24168.

\clearpage
\setcounter{page}{1}
\maketitlesupplementary

\appendix
\renewcommand{\thesection}{\Alph{section}}
\renewcommand{\thefigure}{\Alph{figure}}

\section{Limitations}

\subsection{Quality of Concept Bank}
Our method retrieves explanations from a pre-constructed concept bank; thus, the bank quality and coverage directly limit what can be explained.
If relevant concepts are missing or under-represented, they cannot be retrieved at inference time, potentially yielding incomplete or biased explanations.
This limitation is shared by many concept-based approaches, and improving concept-bank construction is an active research direction.
TEXTER is agnostic to how the bank is built and can naturally benefit from improved construction methods as they emerge.

\par
The number of descriptions per class is also a design choice that trades off coverage and noise.
In practice, low retrieval confidence (e.g., uniformly low similarities or unstable rankings) can indicate insufficient coverage, suggesting that the concept bank should be expanded or adapted to the target domain.

\subsection{Faithfulness evaluation}
Following prior work, we mainly evaluate textual explanations with semantics-based metrics (\cref{sec:quantitative_evaluations}).
These metrics capture semantic consistency, but they do not directly measure causal faithfulness to the classifier's decision; a semantically plausible description may not reflect the evidence actually used by the model.

\par
We also report a perturbation-based evaluation using CLIPSeg masks and class-probability changes (Appendix.~\ref{sec:faithfulness_eval}).
However, this protocol assumes that evidence is well localizable and separable by masking, which can break for fine-grained or distributed cues.
Moreover, when candidate descriptions share largely overlapping regions (e.g., \textit{cat body} vs.\ \textit{striped pattern}), the perturbation signal may be insensitive to the true decision basis.

\par
Overall, robust and widely accepted faithfulness protocols for textual explanations remain underdeveloped, and establishing reliable metrics is an important direction for future work.

\subsection{Computational overhead and dataset dependence}
TEXTER introduces extra cost to train auxiliary modules (SAE/aligner).
This training is offline and one-time (not per-image), but we do not claim efficiency; scalability to larger concept banks and higher-resolution settings remains a limitation.

\par
TEXTER may also be dataset-dependent under distribution shift, as is common for post-hoc methods.
If the concept bank or auxiliary modules are mismatched to the target domain, retrieval and concept-image validity can degrade, making explanations less reliable.
We therefore use the validity check in \cref{sec:eval_concept_image} and treat low validity as a warning signal.

\section{Implementation details}
\label{sec:implementation_details}

\subsection{Details of SAE configuration}
\label{sec:sae_config}
We adopt an overcomplete TopK Sparse Autoencoder~\citep{topk_sae}, 
where the SAE embedding dimensionality exceeds that of the original feature space.
Specifically, 
the embedding dimension (i.e., the dimensionality of $\Psi(f(x))$) 
is set to eight times that of the original feature vector $f(x)$, 
and the Top-K ratio is fixed at $10\,\%$, 
meaning that only the top $10\,\%$ entries 
with the largest magnitudes in each embedding vector are retained.
The SAE is trained following \cref{eq:loss_sae} on the same dataset used to train each classifier (e.g., ImageNet) 
with batch size 1024, learning rate $5\times 10^{-4}$, and the Adam optimizer for 10 epochs.

\subsection{Details of concept bank construction}
\label{sec:detail_concept_bank}
We use an LLM and a VLM to generate the concept bank $\mathcal{B}(x,c)$.
Below, we describe the prompts used for each model.

\par
The LLM is utilized to generate concepts that are generic to class $c$ and not tied to a specific image.
These concepts capture properties commonly associated with the class.
Each concept description is constrained to a short phrase of 1–3 words to encourage compact,
unit-like concepts, 
and descriptions that simply restate the class name are avoided.
For the LLM input, 
we provide the target class name (\{class\_name\}), 
the concepts already generated so far (\{existing\_concepts\}),
and an example question–answer pair, 
and obtain 10 new descriptions in a single inference.
As a post-processing step, duplicate concepts are removed.
Specifically, we use the following prompt:
\begin{promptverb}[title={Prompt template for generating concepts using the LLM},
label={prompt:llm}]
Template variables (filled by the implementation as input variables):
- {class_name}: the target object class name.
- {existing_concepts}: already generated concepts.

Important guidelines for generating visual concepts:
1. Generate GENERAL concepts that can apply to many different photos of the same object type.
2. Include both OBJECT features (e.g., shape, color, parts) AND CONTEXT features (e.g., background, environment, setting).
3. Keep concepts short and specific (1-3 words).
4. DO NOT include class names or object names directly.

Q: What are useful visual features for distinguishing a lemur in a photo?
A: There are several useful visual features to tell there is a lemur in a photo:
- long tail
- large eyes
- gray fur
- trees
- branches
- forest

Q: What are useful features for distinguishing a {class_name} in a photo?
Already generated concepts (DO NOT repeat these): {existing_concepts}.
A: There are several useful visual features to tell there is a {class_name} in a photo. Generate approximately 10 visual concepts to provide comprehensive coverage:
\end{promptverb}

\par
The VLM is employed to generate concepts that are grounded in the visual content of the image for class $c$.
These concepts capture properties that are visually expressed in the specific input image and are therefore complementary to the generic class-level concepts generated by the LLM.
As in the LLM setting, 
each concept description is constrained to a short phrase of 1–3 words, 
and descriptions that simply restate the class name are avoided.
Unlike in the LLM setting, 
we additionally provide the image $x$ as visual input to the VLM,
along with the following text prompt.
The generation procedure and post-processing (e.g., removal of duplicates) are kept consistent with the LLM case.
\begin{promptverb}[title={Prompt template for generating concepts using the VLM},
label={prompt:vlm}]
Template variables (filled by the implementation as input variables):
- {class_name}: the target object class name.
- {existing_concepts}: already generated concepts.

Important guidelines for generating visual concepts:
1. Generate DETAILED and SPECIFIC concepts that can apply to this image.
2. Include both OBJECT features (e.g., shape, color, parts) AND CONTEXT features (e.g., background, environment, setting).
3. Keep concepts short and specific (1-3 words).
4. DO NOT include class names or object names directly.

Examples:
Q: Look at this image carefully. Based on what you can actually see in the image, identify useful visual features that help distinguish this as a koi fish.
A: There are several useful visual features to tell there is a koi fish in a photo:
- bright orange scales
- curved tail fin
- spotted pattern
- long body
- pointed snout
- water surface

Q: Look at this image carefully. Based on what you can actually see in the image, identify useful visual features that help distinguish this as a {class_name}.
Already generated concepts (DO NOT repeat these): {existing_concepts}.
A: There are several useful visual features to tell there is a {class_name} in a photo. Generate approximately 10 visual concepts to provide comprehensive coverage:
\end{promptverb}

\begin{table*}[t]
\centering
\caption{Quantitative evaluation using semantic-based metrics on ImageNet-1K. 
Each metric is computed between the explanations generated by each method and the concept images generated by the proposed method with SAE. 
Best results in bold.
}
\label{tab:results_for_concept_imagenet}
\begin{tabular}{l l c c c c}
\toprule
\textbf{Model} & \textbf{Method} & \textbf{CLIP-Score} $\uparrow$ & \textbf{LPIPS (A)} $\downarrow$ & \textbf{LPIPS (S)} $\downarrow$ & \textbf{FS} $\uparrow$ \\
\midrule
\multirow{3}{*}{ResNet-18}
  & Random                      & 0.2295 & 0.7590 & 0.7065 & 0.6440 \\
  & Text-To-Concept             & 0.2275 & 0.7594 & 0.7054 & 0.6517 \\
  & \cellcolor{gray!12}\mymethod  
                                & \cellcolor{gray!12}\textbf{0.2333} & \cellcolor{gray!12}\textbf{0.7584} & \cellcolor{gray!12}\textbf{0.7036} & \cellcolor{gray!12}\textbf{0.6539} \\
\midrule
\multirow{3}{*}{ResNet-50}
  & Random                      & 0.2306 & 0.7651 & 0.7037 & 0.6814 \\
  & Text-To-Concept             & 0.2309 & 0.7618 & 0.7026 & 0.6856 \\
  & \cellcolor{gray!12}\mymethod
                                & \cellcolor{gray!12}\textbf{0.2376} & \cellcolor{gray!12}\textbf{0.7596} & \cellcolor{gray!12}\textbf{0.7020} & \cellcolor{gray!12}\textbf{0.6961} \\
\midrule
\multirow{3}{*}{DINO ResNet-50}
  & Random                      & 0.2321 & 0.7579 & 0.6944 & 0.5256 \\
  & Text-To-Concept             & 0.2328 & 0.7532 & 0.6907 & 0.5349 \\
  & \cellcolor{gray!12}\mymethod
                                & \cellcolor{gray!12}\textbf{0.2430} & \cellcolor{gray!12}\textbf{0.7506} & \cellcolor{gray!12}\textbf{0.6882} & \cellcolor{gray!12}\textbf{0.5432} \\
\midrule
\multirow{3}{*}{ViT}
  & Random                      & 0.2262 & 0.7736 & 0.6897 & 0.2597 \\
  & Text-To-Concept             & 0.2256 & 0.7746 & 0.6891 & 0.2644 \\
  & \cellcolor{gray!12}\mymethod
                                & \cellcolor{gray!12}\textbf{0.2315} & \cellcolor{gray!12}\textbf{0.7693} & \cellcolor{gray!12}\textbf{0.6846} & \cellcolor{gray!12}\textbf{0.2668} \\
\midrule
\multirow{3}{*}{DINO ViT-S/8}
  & Random                      & 0.2260 & 0.7576 & 0.6915 & 0.4194 \\
  & Text-To-Concept             & 0.2243 & 0.7543 & 0.6898 & 0.4246 \\
  & \cellcolor{gray!12}\mymethod
                                & \cellcolor{gray!12}\textbf{0.2337} & \cellcolor{gray!12}\textbf{0.7541} & \cellcolor{gray!12}\textbf{0.6892} & \cellcolor{gray!12}\textbf{0.4275} \\
\bottomrule
\end{tabular}
\end{table*}

\begin{table*}[t]
\centering
\caption{Quantitative evaluation using semantic-based metrics on ImageNet-1K. 
Each metric is computed between the explanations generated by each method and the original input images. 
Best results in bold.
}
\label{tab:results_for_image_imagenet}
\begin{tabular}{l l c c c c}
\toprule
\textbf{Model} & \textbf{Method} & \textbf{CLIP-Score} $\uparrow$ & \textbf{LPIPS (A)} $\downarrow$ & \textbf{LPIPS (S)} $\downarrow$ & \textbf{FS} $\uparrow$ \\
\midrule
\multirow{3}{*}{ResNet-18}
  & Random                      & 0.3043 & 0.7298 & 0.6180 & 0.7221 \\
  & Text-To-Concept             & \textbf{0.3135} & \textbf{0.7196} & \textbf{0.6099} & \textbf{0.7321} \\
  & \cellcolor{gray!12}\mymethod
                                & \cellcolor{gray!12}0.3077 & \cellcolor{gray!12}0.7278 & \cellcolor{gray!12}0.6191 & \cellcolor{gray!12}0.7255 \\
\midrule
\multirow{3}{*}{ResNet-50}
  & Random                      & 0.3079 & 0.7306 & 0.6231 & 0.7640 \\
  & Text-To-Concept             & \textbf{0.3174} & \textbf{0.7239} & \textbf{0.6144} & \textbf{0.7732} \\
  & \cellcolor{gray!12}\mymethod
                                & \cellcolor{gray!12}0.3108 & \cellcolor{gray!12}0.7302 & \cellcolor{gray!12}0.6219 & \cellcolor{gray!12}0.7683 \\
\midrule
\multirow{3}{*}{DINO ResNet-50}
  & Random                      & 0.3066 & 0.7330 & 0.6230 & 0.6284 \\
  & Text-To-Concept             & \textbf{0.3178} & \textbf{0.7227} & \textbf{0.6139} & \textbf{0.6442} \\
  & \cellcolor{gray!12}\mymethod
                                & \cellcolor{gray!12}0.3098 & \cellcolor{gray!12}0.7270 & \cellcolor{gray!12}0.6210 & \cellcolor{gray!12}0.6316 \\
\midrule
\multirow{3}{*}{ViT}
  & Random                      & 0.3080 & 0.7306 & 0.6219 & 0.5113 \\
  & Text-To-Concept             & \textbf{0.3143} & \textbf{0.7275} & \textbf{0.6160} & \textbf{0.5241} \\
  & \cellcolor{gray!12}\mymethod
                                & \cellcolor{gray!12}0.3113 & \cellcolor{gray!12}0.7318 & \cellcolor{gray!12}0.6237 & \cellcolor{gray!12}0.5157 \\
\midrule
\multirow{3}{*}{DINO ViT-S/8}
  & Random                      & 0.3081 & 0.7290 & 0.6185 & 0.6110 \\
  & Text-To-Concept             & \textbf{0.3184} & \textbf{0.7215} & \textbf{0.6124} & \textbf{0.6363} \\
  & \cellcolor{gray!12}\mymethod
                                & \cellcolor{gray!12}0.3098 & \cellcolor{gray!12}0.7307 & \cellcolor{gray!12}0.6211 & \cellcolor{gray!12}0.6140 \\
\bottomrule
\end{tabular}
\end{table*}

\subsection{Other implementation details}
\label{sec:other_implementation_details}

As described in \cref{sec:vis_concepts}, 
MACO~\citep{maco} is used to generate concept images.
The concept images are generated using 512 iterations following \cref{eq:fv_general}, and all other parameters follow the original paper~\citep{maco}.
Since MACO tends to produce images with repeated patterns, we mitigate the redundancy this may introduce in textual explanation generation by randomly cropping multiple patches from the original concept image.
The side length of each patch is drawn uniformly between 25\,\% and 30\,\% of the original image size,
and the crop center is sampled from a Gaussian around the image center and clipped to remain within the image boundaries.
Each cropped patch is then resized to the original input resolution and mapped into the CLIP feature space via the aligner.
We generate six patches by this process.

\par
For textual explanation generation, 
the similarity between each aligned feature (derived from the six cropped patches) and every candidate in the concept bank $\mathcal{B}(x,c)$ is computed.
For each candidate, the six similarity scores are averaged, and the candidates are sorted according to this averaged score to produce the final ranked list of textual explanations.

\section{Additional quantitative results}
\label{sec:additional_quantitative}

\subsection{Experimental settings for semantic-based metrics}
\label{sec:experimental_settings}
We detail each metric introduced in \cref{sec:quantitative_evaluations}.

\par
For CLIP-Score, 
we use the ViT-B/16 image encoder.
The text prompt is set as
``a photo of \{class\_name\} showing $T$,''
where \{class\_name\} is the target class name and $T$ is the set of generated concepts.
Here, each description in $T$ is separated by a comma.

\par
For computing LPIPS and Feature Similarity, 
we generate an image $x_g$ from $T$ using Stable Diffusion.
We use the ``stable-diffusion-v1-5'' model.
The prompt for Stable Diffusion is formulated similarly to that used for CLIP-Score, 
as
``a \{class\_name\} showing $T$.''

\subsection{Additional results of semantic-based evaluations}
\label{sec:additional_semantic_eval}
We conduct quantitative evaluations using the semantic-based metrics introduced in \cref{sec:quantitative_evaluations} on the ImageNet-1K dataset.
Each metric is computed on 1,000 randomly selected images, consisting of 200 classes with five images per class.

\par
Table~\ref{tab:results_for_concept_imagenet} shows the results comparing the concept images generated by the proposed method with the textual explanations produced by each method.
Across all metrics and models, 
the proposed method consistently outperforms the baselines, 
indicating that it generates explanations with better semantic alignment to the concept images.
Since ImageNet images typically contain a single large object, 
the type of visual features are relatively limited; therefore, the numerical differences appear smaller compared with \cref{tab:results_for_concept_pascal} (evaluated on PASCAL VOC).
Nevertheless, the proposed method consistently achieves superior scores, demonstrating its effectiveness.

\par
As a complementary analysis,
we also evaluate each method by comparing its explanations with the original input images, 
following prior work~\citep{zsnle}.
In this setting, 
higher scores indicate that the explanations describe the overall visual content of the input image well, 
rather than strictly reflecting the model’s decision rationale.
The results are summarized in \cref{tab:results_for_image_imagenet}.
As expected, Text-To-Concept achieves the best performance across all models,
which is consistent with its design: it aligns global image features with text and is intended to describe the input image itself.
These results therefore complement \cref{tab:results_for_concept_pascal,tab:results_for_concept_imagenet}:
while Text-To-Concept is better aligned with the input images,
the proposed method provides explanations that are more semantically aligned with the decision-critical concept images.

\begin{table}[t]
\centering
\caption{Evaluation using segmentation masks under insertion (Ins) and deletion (Del) settings.
Best results are shown in bold.
}
\label{tab:ins_del}
\begin{tabular}{l l c c}
\toprule
\textbf{Model} & \textbf{Method} & \textbf{Ins} $\uparrow$ & \textbf{Del} $\downarrow$  \\
\midrule
\multirow{3}{*}{ResNet-18}
  & Random                      & 0.4690 & 0.2775  \\
  & Text-To-Concept             & 0.4756 & 0.2741  \\
  & \cellcolor{gray!12}\mymethod
                                & \cellcolor{gray!12}\textbf{0.5081} & \cellcolor{gray!12}\textbf{0.2462}  \\
\midrule
\multirow{3}{*}{ResNet-50}
  & Random                      & 0.4822 & 0.2999 \\
  & Text-To-Concept             & 0.4810 & 0.3080  \\
  & \cellcolor{gray!12}\mymethod
                                & \cellcolor{gray!12}\textbf{0.4901} & \cellcolor{gray!12}\textbf{0.2797} \\
\midrule
\multirow{3}{*}{DINO ResNet-50}
  & Random                      & 0.5077 & 0.3383  \\
  & Text-To-Concept             & 0.4977 & 0.3380 \\
  & \cellcolor{gray!12}\mymethod
                                & \cellcolor{gray!12}\textbf{0.5494} & \cellcolor{gray!12}\textbf{0.2880}  \\
\midrule
\multirow{3}{*}{ViT}
  & Random                      & 0.3839 & 0.2916\\
  & Text-To-Concept             & 0.3896 & 0.2922\\
  & \cellcolor{gray!12}\mymethod
                                & \cellcolor{gray!12}\textbf{0.4052} & \cellcolor{gray!12}\textbf{0.2699} \\
\midrule
\multirow{3}{*}{DINO ViT-S/8}
  & Random                      & 0.4319 & 0.3194  \\
  & Text-To-Concept             & 0.4465 & 0.3276  \\
  & \cellcolor{gray!12}\mymethod
                                & \cellcolor{gray!12}\textbf{0.4698} & \cellcolor{gray!12}\textbf{0.2925} \\
\bottomrule
\end{tabular}
\end{table}

\subsection{Perturbation-based faithfulness evaluation}
\label{sec:faithfulness_eval}
To complement semantics-based metrics, 
we additionally evaluate faithfulness with a perturbation test that links each retrieved text explanation to an image region.
For each explanation, we use CLIPSeg~\citep{clip_seg} to obtain a segmentation mask scores,
then perform insertion/deletion by progressively inserting high-score regions into a blank image or deleting them from the original image.
We measure the area under the curve (AUC) of the target-class score change, 
following standard perturbation-based faithfulness protocols.
We run this evaluation under the same setting as \cref{sec:multilabel}.

\par
Table~\ref{tab:ins_del} shows the results.
TEXTER yields consistently higher AUC than baselines, indicating that the regions implied by its retrieved explanations have a larger impact on the target prediction.

\begin{figure*}[t]
  \centering
  \includegraphics[width=\linewidth]{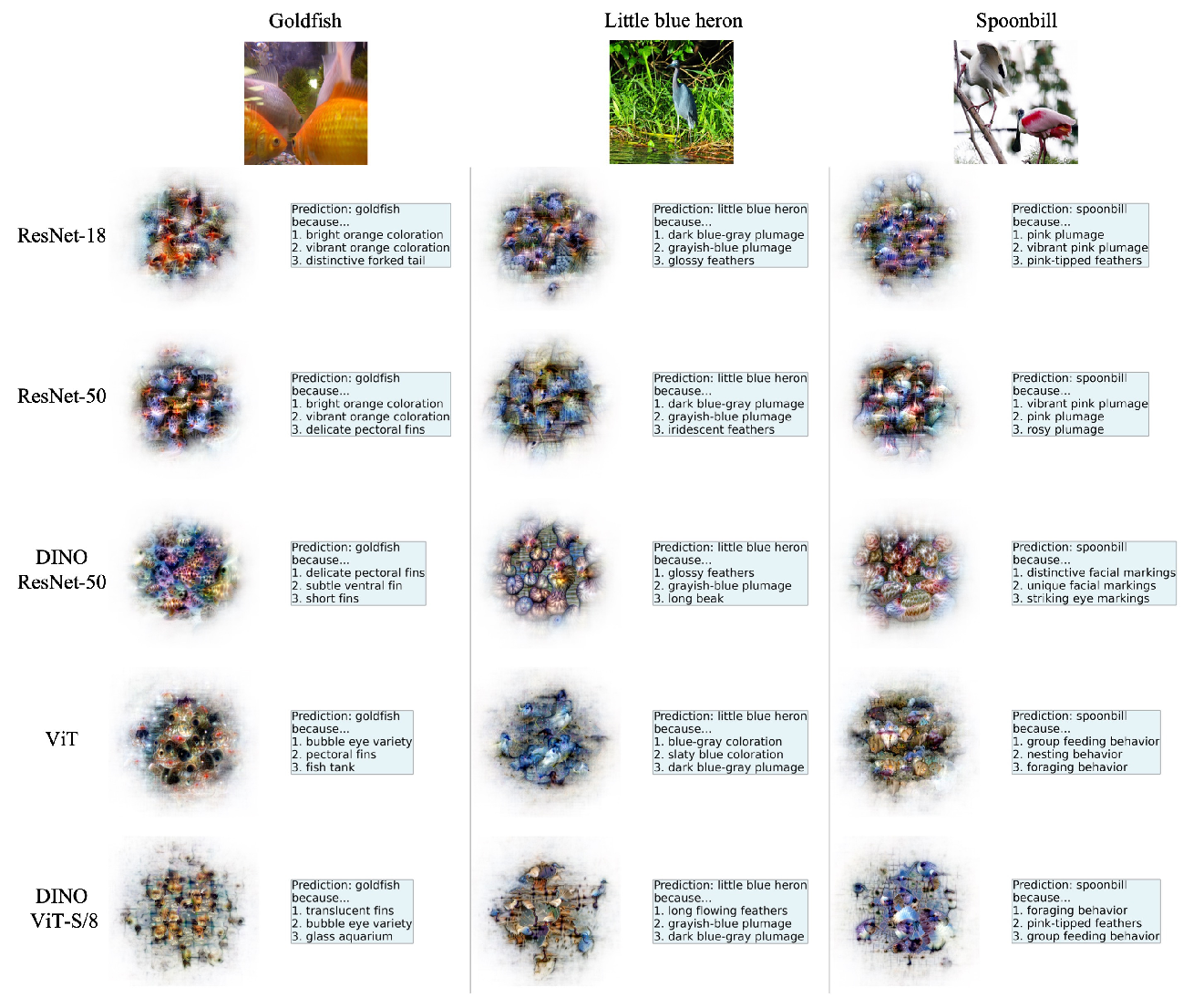}
  \caption{Qualitative comparison of the textual explanations and concept images generated by the proposed method for the same prediction across ResNet-18, ResNet-50, DINO ResNet-50, ViT, and DINO ViT-S/8. Each row corresponds to the results from one model, and each column corresponds to one input image and its predicted class.}
  \label{fig:more_comp1}
\end{figure*}

\begin{figure*}[t]
  \centering
  \includegraphics[width=\linewidth]{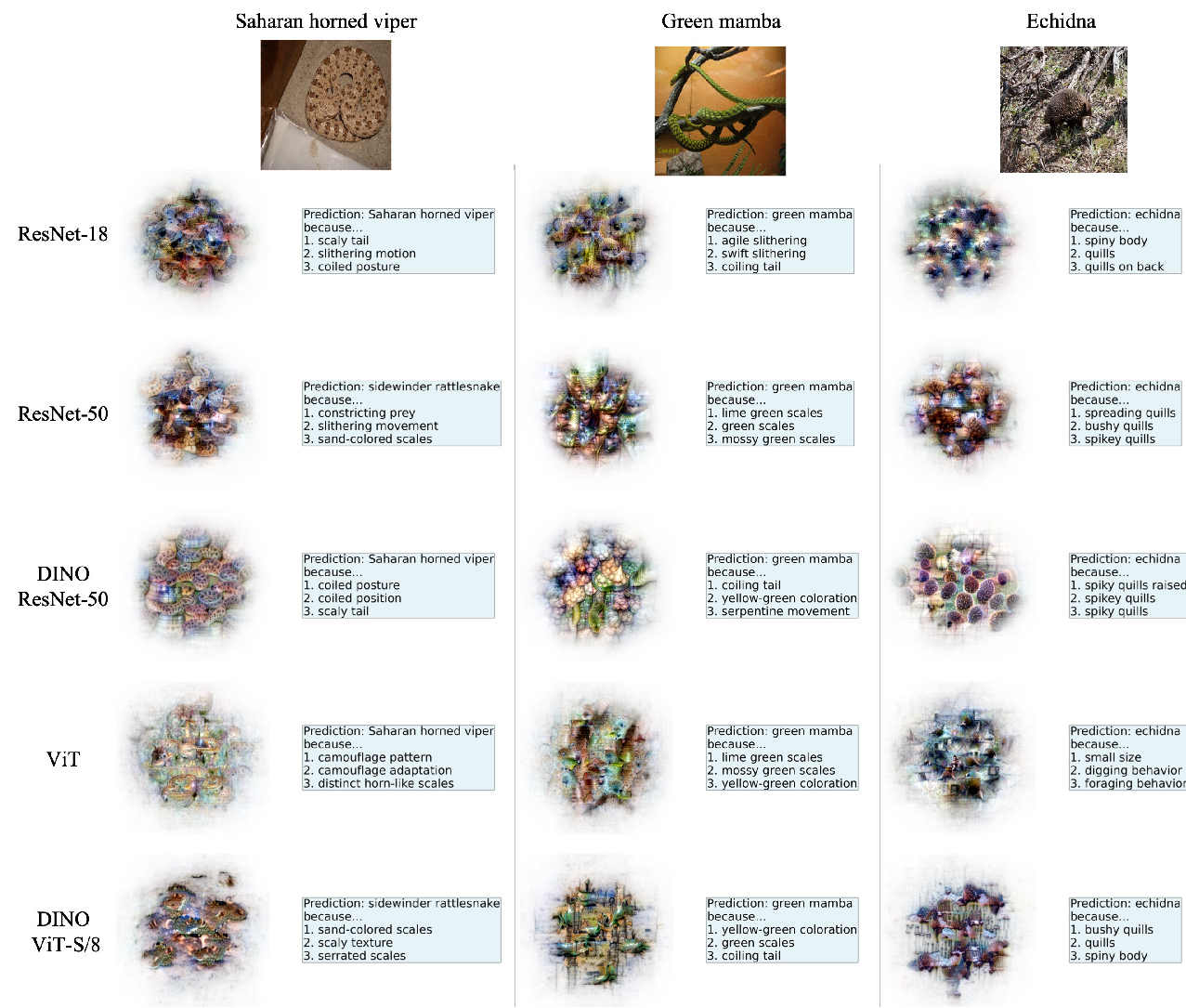}
  \caption{Qualitative comparison of the textual explanations and concept images generated by the proposed method for the same prediction across ResNet-18, ResNet-50, DINO ResNet-50, ViT, and DINO ViT-S/8. Each row corresponds to the results from one model, and each column corresponds to one input image and its predicted class.}
  \label{fig:more_comp2}
\end{figure*}

\section{Additional qualitative results}
\label{sec:additional_qualitative}

\subsection{Additional comparison results across models}
\label{sec:additional_comparison}
\begin{figure*}[t]
  \centering
  \includegraphics[width=\linewidth]{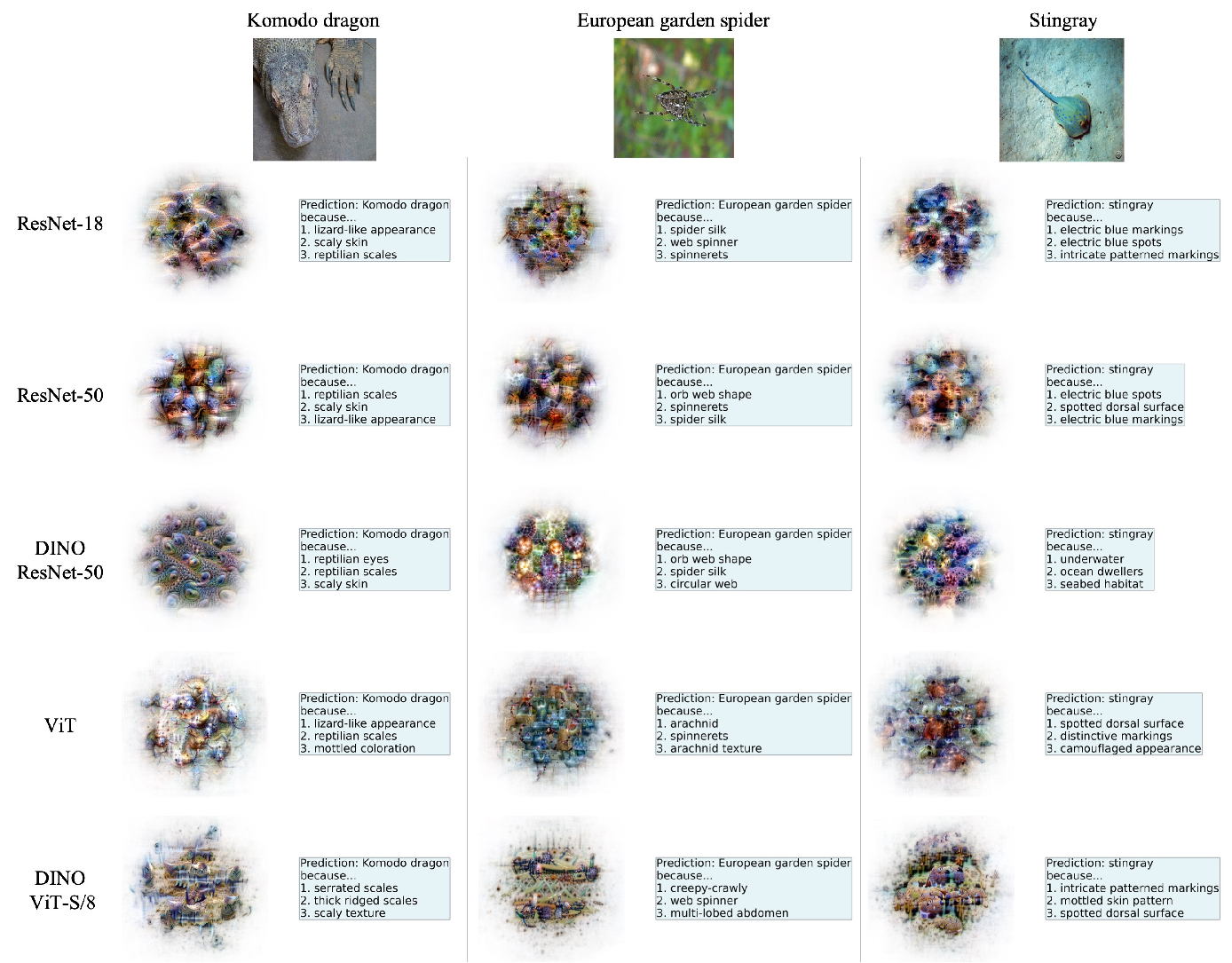}
  \caption{Qualitative comparison of the textual explanations and concept images generated by the proposed method for the same prediction across ResNet-18, ResNet-50, DINO ResNet-50, ViT, and DINO ViT-S/8. Each row corresponds to the results from one model, and each column corresponds to one input image and its predicted class.}
  \label{fig:more_comp3}
\end{figure*}

We show additional comparisons of the generated explanations for the same prediction across models in \cref{fig:more_comp1,fig:more_comp2,fig:more_comp3}.
These examples provide insight into each model’s reasoning.

\par
One example is the difference in the features recognized by CNNs and Transformers.
In particular, Transformers use contextual information in addition to appearance features.
For instance, in the \textit{goldfish} example in \cref{fig:more_comp1},
the CNNs (ResNet-18 and ResNet-50) focus on specific colors such as ``bright orange coloration,''
whereas the Transformer model DINO ViT-S/8 focuses on background cues such as ``glass aquarium.''
Another example is \textit{spoonbill} in \cref{fig:more_comp1}, 
where the Transformer models (ViT and DINO ViT-S/8) capture interactions between two birds, such as ``group feeding behavior.''

\par
We also observe differences in the features used to distinguish similar classes.
For example, as shown for \textit{little blue heron} and \textit{spoonbill} in \cref{fig:more_comp1},
the models often rely on color cues:
the explanations for \textit{little blue heron} include blue-related features such as ``dark blue-gray plumage,''
whereas those for \textit{spoonbill} include pink-related features such as ``pink plumage''.
Another example is \textit{saharan horned viper} and \textit{green mamba} in \cref{fig:more_comp2}.
Here, 
the models appear to distinguish the two classes based on the appearance of the snakes’ scales.
The explanations for \textit{saharan horned viper} include desert-related features, 
such as ``sand-colored scales'' in ResNet-50 and ``camouflage pattern'' in ViT,
whereas those for \textit{green mamba} include green-related features, 
such as ``green scales'' in ResNet-50 and ``yellow-green coloration'' in DINO ViT-S/8.

\par
As shown in these examples,
textual explanations reveal which semantic cues the models rely on.
While such insights are hard to obtain from attribution methods that only highlight important regions as heat maps~\citep{gradcam,score_cam,shap},
the proposed method additionally links each textual explanation to a corresponding concept image,
grounding the text in concrete visual patterns and facilitating a deeper understanding of the model’s behavior.

{
    \small
    \begin{thebibliography}{61}
\providecommand{\natexlab}[1]{#1}
\providecommand{\url}[1]{\texttt{#1}}
\expandafter\ifx\csname urlstyle\endcsname\relax
  \providecommand{\doi}[1]{doi: #1}\else
  \providecommand{\doi}{doi: \begingroup \urlstyle{rm}\Url}\fi

\bibitem[Achiam et~al.()]{openai2024gpt4technicalreport}
Josh Achiam et~al.
\newblock Gpt-4 technical report.
\newblock arXiv: 2303.08774, 2023.

\bibitem[Bai et~al.(2023)Bai, Bai, Yang, Wang, Tan, Wang, Lin, Zhou, and Zhou]{Qwen-VL}
Jinze Bai, Shuai Bai, Shusheng Yang, Shijie Wang, Sinan Tan, Peng Wang, Junyang Lin, Chang Zhou, and Jingren Zhou.
\newblock Qwen-vl: A versatile vision-language model for understanding, localization, text reading, and beyond.
\newblock \emph{arXiv preprint arXiv:2308.12966}, 2023.

\bibitem[Balasubramanian et~al.(2024)Balasubramanian, Basu, and Feizi]{beyond_clip}
Sriram Balasubramanian, Samyadeep Basu, and Soheil Feizi.
\newblock Decomposing and interpreting image representations via text in vits beyond {CLIP}.
\newblock In \emph{Advances in Neural Information Processing Systems (NeurIPS)}, 2024.

\bibitem[Bau et~al.(2017)Bau, Zhou, Khosla, Oliva, and Torralba]{8099837}
David Bau, Bolei Zhou, Aditya Khosla, Aude Oliva, and Antonio Torralba.
\newblock { Network Dissection: Quantifying Interpretability of Deep Visual Representations }.
\newblock In \emph{2017 IEEE Conference on Computer Vision and Pattern Recognition (CVPR)}, pages 3319--3327, 2017.

\bibitem[Bhalla et~al.(2024)Bhalla, Oesterling, Srinivas, Calmon, and Lakkaraju]{bhalla24}
Usha Bhalla, Alex Oesterling, Suraj Srinivas, Flávio~P. Calmon, and Himabindu Lakkaraju.
\newblock Interpreting clip with sparse linear concept embeddings (splice).
\newblock \emph{CoRR}, abs/2402.10376, 2024.

\bibitem[Caron et~al.(2021)Caron, Touvron, Misra, J\'egou, Mairal, Bojanowski, and Joulin]{dino}
Mathilde Caron, Hugo Touvron, Ishan Misra, Herv\'e J\'egou, Julien Mairal, Piotr Bojanowski, and Armand Joulin.
\newblock Emerging properties in self-supervised vision transformers.
\newblock In \emph{Proceedings of the IEEE International Conference on Computer Vision (ICCV)}, 2021.

\bibitem[Dani et~al.(2023)Dani, Rio-Torto, Alaniz, and Akata]{debil}
Meghal Dani, Isabel Rio-Torto, Stephan Alaniz, and Zeynep Akata.
\newblock Devil: Decoding vision features into language.
\newblock \emph{CoRR}, 2023.

\bibitem[Deng et~al.(2009)Deng, Dong, Socher, Li, Li, and Fei-Fei]{imagenet}
Jia Deng, Wei Dong, Richard Socher, Li-Jia Li, Kai Li, and Li Fei-Fei.
\newblock Imagenet: A large-scale hierarchical image database.
\newblock In \emph{Proceedings of the IEEE/CVF Conference on Computer Vision and Pattern Recognition (CVPR)}, pages 248--255, 2009.

\bibitem[Dosovitskiy et~al.(2021)Dosovitskiy, Beyer, Kolesnikov, Weissenborn, Zhai, Unterthiner, Dehghani, Minderer, Heigold, Gelly, Uszkoreit, and Houlsby]{vit}
Alexey Dosovitskiy, Lucas Beyer, Alexander Kolesnikov, Dirk Weissenborn, Xiaohua Zhai, Thomas Unterthiner, Mostafa Dehghani, Matthias Minderer, Georg Heigold, Sylvain Gelly, Jakob Uszkoreit, and Neil Houlsby.
\newblock An image is worth 16x16 words: Transformers for image recognition at scale.
\newblock In \emph{The International Conference on Learning Representations (ICLR)}, 2021.

\bibitem[Everingham et~al.(2010)Everingham, Gool, Williams, Winn, and Zisserman]{voc}
Mark Everingham, Luc~Van Gool, Christopher K.~I. Williams, John~M. Winn, and Andrew Zisserman.
\newblock The pascal visual object classes (voc) challenge.
\newblock \emph{Int. J. Comput. Vis.}, 88\penalty0 (2):\penalty0 303--338, 2010.

\bibitem[FEL et~al.(2023{\natexlab{a}})FEL, Boissin, Boutin, Picard, Novello, Colin, Linsley, ROUSSEAU, Cadene, Goetschalckx, Gardes, and Serre]{maco}
Thomas FEL, Thibaut Boissin, Victor Boutin, Agustin~Martin Picard, Paul Novello, Julien Colin, Drew Linsley, Tom ROUSSEAU, Remi Cadene, Lore Goetschalckx, Laurent Gardes, and Thomas Serre.
\newblock Unlocking feature visualization for deep network with {MA}gnitude constrained optimization.
\newblock In \emph{Advances in Neural Information Processing Systems (NeurIPS)}, 2023{\natexlab{a}}.

\bibitem[FEL et~al.(2023{\natexlab{b}})FEL, Boutin, B{\'e}thune, Cadene, Moayeri, And{\'e}ol, Chalvidal, and Serre]{fel2023a}
Thomas FEL, Victor Boutin, Louis B{\'e}thune, Remi Cadene, Mazda Moayeri, L{\'e}o And{\'e}ol, Mathieu Chalvidal, and Thomas Serre.
\newblock A holistic approach to unifying automatic concept extraction and concept importance estimation.
\newblock In \emph{Thirty-seventh Conference on Neural Information Processing Systems}, 2023{\natexlab{b}}.

\bibitem[Fel et~al.(2023)Fel, Picard, Bethune, Boissin, Vigouroux, Colin, Cadène, and Serre]{fel2023}
Thomas Fel, Agustin Picard, Louis Bethune, Thibaut Boissin, David Vigouroux, Julien Colin, Rémi Cadène, and Thomas Serre.
\newblock Craft: Concept recursive activation factorization for explainability, 2023.

\bibitem[Gandelsman et~al.(2024)Gandelsman, Efros, and Steinhardt]{interpreting_clip}
Yossi Gandelsman, Alexei~A Efros, and Jacob Steinhardt.
\newblock Interpreting {CLIP}'s image representation via text-based decomposition.
\newblock In \emph{The Twelfth International Conference on Learning Representations}, 2024.

\bibitem[Gao et~al.(2025)Gao, la~Tour, Tillman, Goh, Troll, Radford, Sutskever, Leike, and Wu]{topk_sae}
Leo Gao, Tom~Dupre la Tour, Henk Tillman, Gabriel Goh, Rajan Troll, Alec Radford, Ilya Sutskever, Jan Leike, and Jeffrey Wu.
\newblock Scaling and evaluating sparse autoencoders.
\newblock In \emph{The International Conference on Learning Representations (ICLR)}, 2025.

\bibitem[Gorgun et~al.(2025)Gorgun, Schiele, and Fischer]{vital}
Ada Gorgun, Bernt Schiele, and Jonas Fischer.
\newblock Vital: More understandable feature visualization through distribution alignment and relevant information flow, 2025.

\bibitem[He et~al.(2016)He, Zhang, Ren, and Sun]{resnet}
Kaiming He, Xiangyu Zhang, Shaoqing Ren, and Jian Sun.
\newblock {Deep Residual Learning for Image Recognition}.
\newblock In \emph{Proceedings of the IEEE/CVF Conference on Computer Vision and Pattern Recognition (CVPR)}, pages 770--778, 2016.

\bibitem[Hendricks et~al.(2016)Hendricks, Akata, Rohrbach, Donahue, Schiele, and Darrell]{generate_visual_explanation}
Lisa~Anne Hendricks, Zeynep Akata, Marcus Rohrbach, Jeff Donahue, Bernt Schiele, and Trevor Darrell.
\newblock Generating visual explanations.
\newblock In \emph{Proceedings of the European Conference on Computer Vision (ECCV)}, pages 3--19, 2016.

\bibitem[Hernandez et~al.(2022)Hernandez, Schwettmann, Bau, Bagashvili, Torralba, and Andreas]{milan}
Evan Hernandez, Sarah Schwettmann, David Bau, Teona Bagashvili, Antonio Torralba, and Jacob Andreas.
\newblock Natural language descriptions of deep visual features.
\newblock In \emph{The International Conference on Learning Representations (ICLR)}, 2022.

\bibitem[Hessel et~al.(2021)Hessel, Holtzman, Forbes, Le~Bras, and Choi]{clip_score}
Jack Hessel, Ari Holtzman, Maxwell Forbes, Ronan Le~Bras, and Yejin Choi.
\newblock {CLIPS}core: A reference-free evaluation metric for image captioning.
\newblock In \emph{Proceedings of the 2021 Conference on Empirical Methods in Natural Language Processing}, pages 7514--7528, 2021.

\bibitem[Iandola et~al.(2016)Iandola, Han, Moskewicz, Ashraf, Dally, and Keutzer]{squeezenet}
Forrest~N. Iandola, Song Han, Matthew~W. Moskewicz, Khalid Ashraf, William~J. Dally, and Kurt Keutzer.
\newblock Squeezenet: Alexnet-level accuracy with 50x fewer parameters and <0.5mb model size, 2016.

\bibitem[Jiang et~al.(2024)Jiang, Khorram, and Fuxin]{decision_making}
Mingqi Jiang, Saeed Khorram, and Li Fuxin.
\newblock Comparing the decision-making mechanisms by transformers and cnns via explanation methods.
\newblock In \emph{Proceedings of the IEEE Conference on Computer Vision and Pattern Recognition (CVPR)}, pages 9546--9555, 2024.

\bibitem[Kim et~al.(2018)Kim, Wattenberg, Gilmer, Cai, Wexler, Viégas, and Sayres]{tcav}
Been Kim, Martin Wattenberg, Justin Gilmer, Carrie~J. Cai, James Wexler, Fernanda~B. Viégas, and Rory Sayres.
\newblock Interpretability beyond feature attribution: Quantitative testing with concept activation vectors (tcav).
\newblock In \emph{Proceedings of the International Conference on Machine Learning (ICML)}, pages 2673--2682, 2018.

\bibitem[Koh et~al.(2020)Koh, Nguyen, Tang, Mussmann, Pierson, Kim, and Liang]{cbn}
Pang~Wei Koh, Thao Nguyen, Yew~Siang Tang, Stephen Mussmann, Emma Pierson, Been Kim, and Percy Liang.
\newblock Concept bottleneck models.
\newblock In \emph{Proceedings of the International Conference on Machine Learning (ICML)}, 2020.

\bibitem[Krizhevsky et~al.(2012)Krizhevsky, Sutskever, and Hinton]{alexnet}
Alex Krizhevsky, Ilya Sutskever, and Geoffrey~E Hinton.
\newblock Imagenet classification with deep convolutional neural networks.
\newblock In \emph{Advances in Neural Information Processing Systems (NeurIPS)}, 2012.

\bibitem[Li et~al.(2022)Li, Li, Xiong, and Hoi]{pmlr-v162-li22n}
Junnan Li, Dongxu Li, Caiming Xiong, and Steven Hoi.
\newblock {BLIP}: Bootstrapping language-image pre-training for unified vision-language understanding and generation.
\newblock In \emph{Proceedings of the 39th International Conference on Machine Learning}, pages 12888--12900, 2022.

\bibitem[Liu et~al.(2023)Liu, Li, Wu, and Lee]{NEURIPS2023_6dcf277e}
Haotian Liu, Chunyuan Li, Qingyang Wu, and Yong~Jae Lee.
\newblock Visual instruction tuning.
\newblock In \emph{Advances in Neural Information Processing Systems}, pages 34892--34916, 2023.

\bibitem[Liu et~al.(2025)Liu, Zhang, and Gu]{hybrid_cbn}
Yang Liu, Tianwei Zhang, and Shi Gu.
\newblock Hybrid concept bottleneck models.
\newblock In \emph{Proceedings of the IEEE/CVF Conference on Computer Vision and Pattern Recognition (CVPR)}, pages 20179--20189, 2025.

\bibitem[L\"uddecke and Ecker(2022)]{clip_seg}
Timo L\"uddecke and Alexander Ecker.
\newblock Image segmentation using text and image prompts.
\newblock In \emph{Proceedings of the IEEE/CVF Conference on Computer Vision and Pattern Recognition (CVPR)}, pages 7086--7096, 2022.

\bibitem[Lundberg and Lee(2017)]{shap}
Scott~M. Lundberg and Su-In Lee.
\newblock A unified approach to interpreting model predictions.
\newblock In \emph{Advances in Neural Information Processing Systems (NeurIPS)}, page 4768–4777, 2017.

\bibitem[Menon and Vondrick(2023)]{menon}
Sachit Menon and Carl Vondrick.
\newblock Visual classification via description from large language models, 2023.

\bibitem[Moayeri et~al.(2023)Moayeri, Rezaei, Sanjabi, and Feizi]{text_to_concept}
Mazda Moayeri, Keivan Rezaei, Maziar Sanjabi, and Soheil Feizi.
\newblock Text-to-concept (and back) via cross-model alignment.
\newblock In \emph{Proceedings of the International Conference on Machine Learning (ICML)}, 2023.

\bibitem[Nguyen et~al.(2016)Nguyen, Dosovitskiy, Yosinski, Brox, and Clune]{nguyen16}
Anh Nguyen, Alexey Dosovitskiy, Jason Yosinski, Thomas Brox, and Jeff Clune.
\newblock Synthesizing the preferred inputs for neurons in neural networks via deep generator networks.
\newblock In \emph{Advances in Neural Information Processing Systems (NeurIPS)}, page 3395–3403, 2016.

\bibitem[Nguyen et~al.(2017)Nguyen, Clune, Bengio, Dosovitskiy, and Yosinski]{nguyen17}
Anh Nguyen, Jeff Clune, Y. Bengio, Alexey Dosovitskiy, and Jason Yosinski.
\newblock Plug \& play generative networks: Conditional iterative generation of images in latent space.
\newblock In \emph{Proceedings of the IEEE/CVF Conference on Computer Vision and Pattern Recognition (CVPR)}, pages 3510--3520, 2017.

\bibitem[Oikarinen et~al.(2023)Oikarinen, Das, Nguyen, and Weng]{label_free_cbn}
Tuomas Oikarinen, Subhro Das, Lam~M. Nguyen, and Tsui-Wei Weng.
\newblock Label-free concept bottleneck models.
\newblock In \emph{The International Conference on Learning Representations (ICLR)}, 2023.

\bibitem[Oikarinen and Weng(2023)]{clip_dissect}
Tuomas~P. Oikarinen and Tsui-Wei Weng.
\newblock Clip-dissect: Automatic description of neuron representations in deep vision networks.
\newblock In \emph{The International Conference on Learning Representations (ICLR)}, 2023.

\bibitem[Olah et~al.(2017)Olah, Schubert, and Mordvintsev]{fv}
Christopher Olah, Ludwig Schubert, and Alexander Mordvintsev.
\newblock Feature visualization.
\newblock \emph{Distill}, 2017.

\bibitem[OpenAI(2025)]{gpt3.5}
OpenAI.
\newblock Gpt-3.5 turbo models.
\newblock \url{https://platform.openai.com/docs/models/gpt-3-5}, 2025.
\newblock Accessed 2025-10-13.

\bibitem[Park et~al.(2018)Park, Hendricks, Akata, Rohrbach, Schiele, Darrell, and Rohrbach]{park_nle}
Dong~Huk Park, Lisa~Anne Hendricks, Zeynep Akata, Anna Rohrbach, Bernt Schiele, Trevor Darrell, and Marcus Rohrbach.
\newblock Multimodal explanations: Justifying decisions and pointing to the evidence.
\newblock In \emph{Proceedings of the IEEE Conference on Computer Vision and Pattern Recognition (CVPR)}, 2018.

\bibitem[Radford et~al.(2019)Radford, Wu, Child, Luan, Amodei, and Sutskever]{gpt2}
Alec Radford, Jeffrey Wu, Rewon Child, David Luan, Dario Amodei, and Ilya Sutskever.
\newblock Language models are unsupervised multitask learners.
\newblock \emph{OpenAI}, 2019.

\bibitem[Radford et~al.(2021)Radford, Kim, Hallacy, Ramesh, Goh, Agarwal, Sastry, Askell, Mishkin, Clark, Krueger, and Sutskever]{clip}
Alec Radford, Jong~Wook Kim, Chris Hallacy, Aditya Ramesh, Gabriel Goh, Sandhini Agarwal, Girish Sastry, Amanda Askell, Pamela Mishkin, Jack Clark, Gretchen Krueger, and Ilya Sutskever.
\newblock Learning transferable visual models from natural language supervision.
\newblock In \emph{Proceedings of the International Conference on Machine Learning (ICML)}, pages 8748--8763, 2021.

\bibitem[Raghu et~al.(2021)Raghu, Unterthiner, Kornblith, Zhang, and Dosovitskiy]{vit_cnn1}
Maithra Raghu, Thomas Unterthiner, Simon Kornblith, Chiyuan Zhang, and Alexey Dosovitskiy.
\newblock Do vision transformers see like convolutional neural networks?
\newblock In \emph{Advances in Neural Information Processing Systems (NeurIPS)}, pages 12116--12128, 2021.

\bibitem[Rombach et~al.(2022)Rombach, Blattmann, Lorenz, Esser, and Ommer]{stable_diffusion}
Robin Rombach, Andreas Blattmann, Dominik Lorenz, Patrick Esser, and Bj\"orn Ommer.
\newblock High-resolution image synthesis with latent diffusion models.
\newblock In \emph{Proceedings of the IEEE/CVF Conference on Computer Vision and Pattern Recognition (CVPR)}, pages 10684--10695, 2022.

\bibitem[Salewski et~al.(2024)Salewski, Koepke, Lensch, and Akata]{zs-a2t}
Leonard Salewski, A.~Sophia Koepke, Hendrik P.~A. Lensch, and Zeynep Akata.
\newblock Zero-shot translation of attention patterns in vqa models to natural language.
\newblock In \emph{Pattern Recognition}, pages 378--393, Cham, 2024.

\bibitem[Sammani and Deligiannis(2023)]{uni_nlx}
Fawaz Sammani and Nikos Deligiannis.
\newblock Uni-nlx: Unifying textual explanations for vision and vision-language tasks.
\newblock In \emph{Proceedings of the IEEE/CVF International Conference on Computer Vision (ICCV) Workshops}, pages 4634--4639, 2023.

\bibitem[Sammani and Deligiannis(2025)]{zsnle}
Fawaz Sammani and Nikos Deligiannis.
\newblock Zero-shot natural language explanations.
\newblock In \emph{The International Conference on Learning Representations (ICLR)}, 2025.

\bibitem[Sammani et~al.(2022)Sammani, Mukherjee, and Deligiannis]{nlx_gpt}
Fawaz Sammani, Tanmoy Mukherjee, and Nikos Deligiannis.
\newblock Nlx-gpt: A model for natural language explanations in vision and vision-language tasks.
\newblock In \emph{Proceedings of the IEEE/CVF Conference on Computer Vision and Pattern Recognition (CVPR)}, pages 8322--8332, 2022.

\bibitem[Selvaraju et~al.(2017)Selvaraju, Cogswell, Das, Vedantam, Parikh, and Batra]{gradcam}
Ramprasaath~R. Selvaraju, Michael Cogswell, Abhishek Das, Ramakrishna Vedantam, Devi Parikh, and Dhruv Batra.
\newblock Grad-cam: Visual explanations from deep networks via gradient-based localization.
\newblock In \emph{Proceedings of the IEEE International Conference on Computer Vision (ICCV)}, pages 618--626, 2017.

\bibitem[Shang et~al.(2024)Shang, Zhou, Zhang, Ni, Yang, and Wang]{incremental_cbn}
Chenming Shang, Shiji Zhou, Hengyuan Zhang, Xinzhe Ni, Yujiu Yang, and Yuwang Wang.
\newblock Incremental residual concept bottleneck models.
\newblock In \emph{Proceedings of the IEEE/CVF Conference on Computer Vision and Pattern Recognition (CVPR)}, pages 11030--11040, 2024.

\bibitem[Sharma et~al.(2018)Sharma, Ding, Goodman, and Soricut]{sharma}
Piyush Sharma, Nan Ding, Sebastian Goodman, and Radu Soricut.
\newblock Conceptual captions: A cleaned, hypernymed, image alt-text dataset for automatic image captioning.
\newblock In \emph{Proceedings of the 56th Annual Meeting of the Association for Computational Linguistics (Volume 1: Long Papers)}, pages 2556--2565, 2018.

\bibitem[Shtedritski et~al.(2023)Shtedritski, Rupprecht, and Vedaldi]{shtedritski}
Aleksandar Shtedritski, C. Rupprecht, and Andrea Vedaldi.
\newblock What does clip know about a red circle? visual prompt engineering for vlms, 2023.

\bibitem[Sundararajan et~al.(2017)Sundararajan, Taly, and Yan]{ig}
Mukund Sundararajan, Ankur Taly, and Qiqi Yan.
\newblock Axiomatic attribution for deep networks.
\newblock In \emph{Proceedings of the International Conference on Machine Learning (ICML)}, page 3319–3328, 2017.

\bibitem[Team(2025)]{qwen2.5-VL}
Qwen Team.
\newblock Qwen2.5-vl, 2025.

\bibitem[Thasarathan et~al.(2025)Thasarathan, Forsyth, Fel, Kowal, and Derpanis]{usae}
Harrish Thasarathan, Julian Forsyth, Thomas Fel, Matthew Kowal, and Konstantinos~G. Derpanis.
\newblock Universal sparse autoencoders: Interpretable cross-model concept alignment.
\newblock In \emph{Proceedings of the International Conference on Machine Learning (ICML)}, 2025.

\bibitem[Tuli et~al.(2021)Tuli, Dasgupta, Grant, and Griffiths]{vit_cnn2}
Shikhar Tuli, Ishita Dasgupta, Erin Grant, and Thomas~L. Griffiths.
\newblock Are convolutional neural networks or transformers more like human vision?
\newblock \emph{ArXiv}, abs/2105.07197, 2021.

\bibitem[Wang et~al.(2023)Wang, Li, Nakashima, and Nagahara]{language_cbn}
Bowen Wang, Liangzhi Li, Yuta Nakashima, and Hajime Nagahara.
\newblock Learning bottleneck concepts in image classification.
\newblock In \emph{Proceedings of the IEEE/CVF Conference on Computer Vision and Pattern Recognition (CVPR)}, pages 10962--10971, 2023.

\bibitem[Wang et~al.(2020)Wang, Wang, Du, Yang, Zhang, Ding, Mardziel, and Hu]{score_cam}
Haofan Wang, Zifan Wang, Mengnan Du, Fan Yang, Zijian Zhang, Sirui Ding, Piotr Mardziel, and Xia Hu.
\newblock Score-cam: Score-weighted visual explanations for convolutional neural networks.
\newblock In \emph{Proceedings of the IEEE Conference on Computer Vision and Pattern Recognition (CVPR) Workshops}, pages 111--119, 2020.

\bibitem[Wang et~al.(2024)Wang, Bai, Tan, Wang, Fan, Bai, Chen, Liu, Wang, Ge, Fan, Dang, Du, Ren, Men, Liu, Zhou, Zhou, and Lin]{Qwen2VL}
Peng Wang, Shuai Bai, Sinan Tan, Shijie Wang, Zhihao Fan, Jinze Bai, Keqin Chen, Xuejing Liu, Jialin Wang, Wenbin Ge, Yang Fan, Kai Dang, Mengfei Du, Xuancheng Ren, Rui Men, Dayiheng Liu, Chang Zhou, Jingren Zhou, and Junyang Lin.
\newblock Qwen2-vl: Enhancing vision-language model's perception of the world at any resolution.
\newblock \emph{arXiv preprint arXiv:2409.12191}, 2024.

\bibitem[Yang et~al.(2023)Yang, Panagopoulou, Zhou, Jin, Callison-Burch, and Yatskar]{labo}
Yue Yang, Artemis Panagopoulou, Shenghao Zhou, Daniel Jin, Chris Callison-Burch, and Mark Yatskar.
\newblock Language in a bottle: Language model guided concept bottlenecks for interpretable image classification.
\newblock In \emph{Proceedings of the IEEE/CVF Conference on Computer Vision and Pattern Recognition (CVPR)}, pages 19187--19197, 2023.

\bibitem[Zhang et~al.(2018)Zhang, Isola, Efros, Shechtman, and Wang]{lpips}
Richard Zhang, Phillip Isola, Alexei~A Efros, Eli Shechtman, and Oliver Wang.
\newblock The unreasonable effectiveness of deep features as a perceptual metric.
\newblock In \emph{Proceedings of the IEEE/CVF Conference on Computer Vision and Pattern Recognition (CVPR)}, pages 586--595, 2018.

\bibitem[Zhang et~al.(2021)Zhang, Madumal, Miller, Ehinger, and Rubinstein]{zhang}
Ruihan Zhang, Prashan Madumal, Tim Miller, Krista~A. Ehinger, and Benjamin I.~P. Rubinstein.
\newblock Invertible concept-based explanations for {CNN} models with non-negative concept activation vectors.
\newblock In \emph{Proceedings of the {AAAI} Conference on Artificial Intelligence}, 2021.

\end{thebibliography}

}
\end{document}